석 사 학 위 논 문

M.S. Dissertation

# 각도 차이를 고려한 휴리스틱 함수와 인체 보행의 에너지 근사 함수를 이용한 실시간 A* 적응형 행동 집합 발자국 계획

Real time A* Adaptive Action Set Footstep Planning with Human Locomotion Energy Approximations Considering Angle Difference for Heuristic Function

2019

김 준 하 (金 俊 河 Kim, Joon-Ha)

한 국 과 학 기 술 원

Korea Advanced Institute of Science and Technology

석 사 학 위 논 문

각도 차이를 고려한 휴리스틱 함수와

인체 보행의 에너지 근사 함수를 이용한

실시간 A* 적응형 행동 집합 발자국 계획

2019

김 준 하

한 국 과 학 기 술 원

기계공학과

# 각도 차이를 고려한 휴리스틱 함수와

# 인체 보행의 에너지 근사 함수를 이용한

# 실시간 A* 적응형 행동 집합 발자국 계획

김 준 하

위 논문은 한국과학기술원 석사학위논문으로
학위논문 심사위원회의 심사를 통과하였음

2018년 12월 26일

심사위원장   오 준 호   (인)

심 사 위 원   김 아 영   (인)

심 사 위 원   김 진 환   (인)

# Real time A* Adaptive Action Set Footstep Planning with Human Locomotion Energy Approximations Considering Angle Difference for Heuristic Function

Joon-Ha Kim

Advisor: Jun-Ho Oh

A dissertation submitted to the faculty of
Korea Advanced Institute of Science and Technology in
partial fulfillment of the requirements for the degree of
Master of Science in Mechanical Engineering

Daejeon, Korea
December 26, 2018

Approved by

_______________________
Jun-Ho Oh
Professor of Mechanical Engineering

The study was conducted in accordance with Code of Research Ethics[1].

---

1) Declaration of Ethical Conduct in Research**:** I, as a graduate student of Korea Advanced Institute of Science and Technology, hereby declare that I have not committed any act that may damage the credibility of my research. This includes, but is not limited to, falsification, thesis written by someone else, distortion of research findings, and plagiarism. I confirm that my dissertation contains honest conclusions based on my own careful research under the guidance of my advisor.

초 록

2족 보행 로봇이 다양한 환경에서 원하는 목적지까지 이동하는 네비게이션 문제는 굉장히 중요하다. 그러나 자유도가 높은 2족 보행 로봇의 특성상 연산 시간이 굉장히 오래 걸리기 때문에 실시간으로 네비게이션 문제를 푸는 것은 매우 어려운 일이다. 이를 극복하고자 많은 학자들이 제시한 방법이 발자국 계획을 통한 네비게이션이다. 기존의 발자국 계획들은 A* 알고리즘을 기반으로 목적함수를 단순히 최단 거리나 각도들을 활용한 값을 사용하였으나, 최근 들어 인체 역학 분야에서 널리 쓰이는 인간의 보행 시 소요되는 에너지를 다항 함수로 근사 하여 이를 목적함수화해서 사용하는 방안들이 제시되었다. 또한 실시간 네비게이션을 위하여 A* 알고리즘의 행동 집합을 고정하지 않고 유동적으로 상황에 맞게 개수를 변화시켜가며 사용하여, 연산 시간은 많이 늘리지 않으며 외부 환경과의 충돌 고려는 더 정확히 할 수 있는 방법들이 제시되었다. 본 학위논문에서는 인간의 보행 시 소요되는 에너지를 근사한 다항 함수를 목적함수로 채택하였으며, 기존 연구들에서는 제시되지 않은 로봇과 목적지와의 각도 차이를 고려한 휴리스틱 함수를 새로 제시하였으며, 그 타당성을 수학적으로 증명하였다. 또한 적응형 행동 집합과 인간 보행 관련 에너지를 통합하는 방법을 새로 제안하고자 하며, 이 두가지 특성을 모두 가진 체 효율적인 충돌회피 방법과 극소점 문제를 줄이는 방법도 제시하고자 한다. 이후 이 모든 특징을 담은 발자국 계획 알고리즘을 매핑 알고리즘과 보행 알고리즘에 통합하여 시뮬레이션 및 실제 로봇으로 네비게이션 문제를 풀고자 한다.

핵 심 낱 말   네비게이션, 발자국 계획, 휴머노이드 로봇, 휴리스틱 함수, 인체 보행 에너지 근사 함수, 적응형 행동 집합, 극소점 문제, 충돌 회피.

Abstract

The problem of navigating a bipedal robot to a desired destination in various environments is very important. However, it is very difficult to solve the navigation problem in real time because the computation time is very long due to the nature of the biped robot having a high degree of freedom. In order to overcome this, many scientists suggested navigation through the footstep planning. Usually footstep planning use the shortest distance or angles as the objective function based on the A * algorithm. Recently, the energy required for human walking, which is widely used in human dynamics, approximated by a polynomial function is proposed as a better cost function that explains the bipedal robot's movement. In addition, for the real time navigation, using the action set of the A * algorithm not fixed, but the number changing according to the situation, so that the computation time does not increase much and the methods of considering the collision with the external environment are suggested as a practical method. In this thesis, polynomial function approximating the energy required for human walking is adopted as a cost function, and heuristic function considering the angular difference between the robot and the destination which is not shown in the previous studies is newly proposed and proved. In addition, a new method to integrate the adaptive behavior set and energy related to human walking is proposed. Furthermore, efficient collision avoidance method and a method to reduce the local minimum problem is proposed in this framework. Finally, footstep planning algorithm with all of these features into the mapping algorithm and the walking algorithm to solve the navigation problem is validated with simulation and real robot.

Keywords Navigation, Footstep planning, Humanoid robot, Heuristic function, Human inspired energy approximation, Adaptive action set, Local minimum problem, Collision avoidance

# Table of Contents

# List of Tables

# List of Figures

# Chapter 1. Introduction

## 1.1. Motivation and Brief History of Footstep Planning

Since development of humanoid robots, it has been used in various fields by utilizing its many DOF (degree of freedom) system. Especially in terms of traversability, it has a big advantage for availability to step over or step on an obstacle, than traditional wheeled robots. However, in the same reason, having complex system makes motion generation and control much more complicated. Furthermore, navigation problem, which moves humanoid from point to point with given environments, has been viewed as a major challenge in terms of computational complexity.

In this reason, at 2001 James Kuffner proposed a simplified navigation method using the hybrid dynamics property of a humanoid from a discrete change at every footstep, which plans only the footholds and named it as "Footstep Planning" [1]. In this method, footsteps are planned at a reduced search space first and then, sufficient COM trajectory planning and walking control is done. Including Kuffner, Chestnutt(2003,2005) [2][3], Gutmann(2005) [4], Perin(2011) [5], Hournung(2012) [6], Stumpf(2014) [7], Kanoulas(2016) [8] used discrete search based navigation algorithms like A*[9] and RRT[10]. These methods' complexity quickly increases with the size of the transition model, in this reason above methods has fixed step models and stepping capabilities are limited. However, in real time application to the real robot it still has computational advantage than fully continuous methods like Kanoun (2011) [12], Deits (2014) [13]. So in this thesis, A* is selected as a base algorithm which has more optimality than RRT and computationally faster than continuous methods.

Before moving on, brief explanation of A* algorithm and A* based footstep planning will be done.

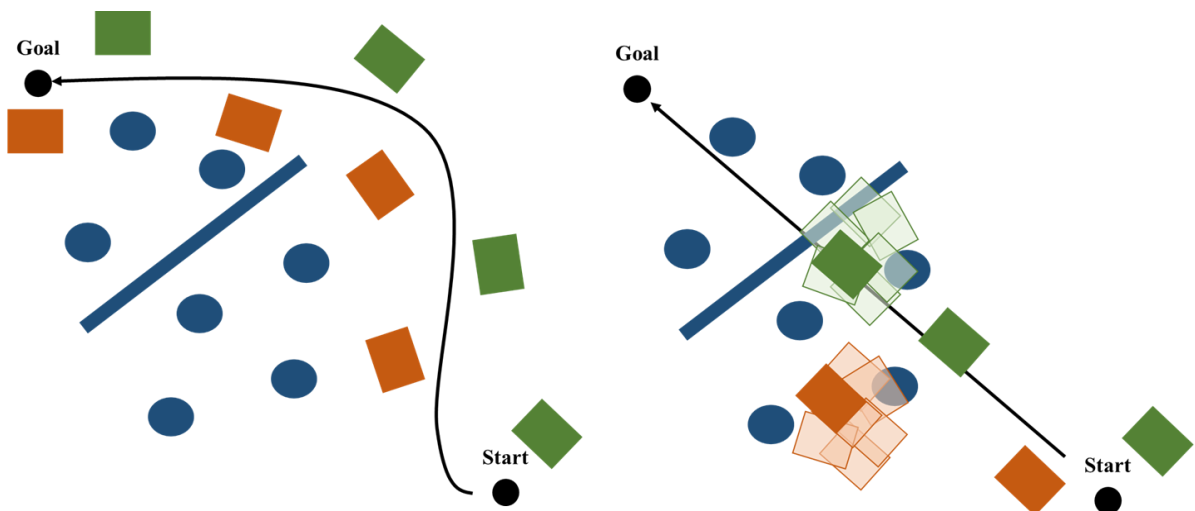

Figure 1.1 Body based navigation and Footstep based navigation

## 1.2. Fundamentals

A* algorithm was first introduced as part of robot Shakey project [9]. Since development, itself or various variants has been used in many path-planning problems. In this section, brief explanation of A* algorithm and its 2d example will be introduced and also Footstep planning method which is based on A* algorithm will be explained.

### 1.2.1. A* algorithm

A* algorithm starts from a 'start node' and aims to find a path to the given 'goal node' having the smallest cost. Cost can be defined as minimal distance or smallest time or energy functions, etc. It finds a path by a sequential order while maintaining a tree of states 's' originating from start node by expanding fixed action pairs 'a' at one of the tree's edge. A* algorithm selects one edge at a time which minimizes priority function 'f(s)', which is sum of cost to come 'g(s)' and heuristic cost to go 'h(s)', until it meets the termination condition. Cost to come g(s) is a summation of each action costs defined by the user to reach the state s from start node. It can also be represented as Equation (1), where $g(s_{before})$ is cost to come of the state right above in the tree and $c(s_{before}, s)$ is cost needed to extend to state s from state $s_{before}$ by certain action. Heuristic cost to go h(s) is an estimation of the cheapest path from state s to goal node. As proven in [11], if the heuristic cost to go h(s) is admissible (never overestimates the actual cost to the goal like Equation (3), A* algorithm gives the optimal path about given action sets and given costs.

$$f(s) = g(s) + h(s) \tag{1}$$

$$g(s) = g(s_{before}) + c(s_{before}, s) \tag{2}$$

$$h(s) \leq \text{actual cost} \tag{3}$$

For example, in a 2d grid space, start node, goal node, state s, action a, g(s), h(s), f(s) can be represented like Figure 1.2. In this example, each action cost $c(s_{before}, s)$ is equally 1 and if it goes to an obstacle (orange: weak, red: strong) it gets additional penalty cost (orange: 1, red: 2). Cost to come g(s) in this example is 2 because it came from the start node by 2 actions. Heuristic cost h(s) is 6 because it can reach to the goal in 6 actions if there is no obstacle. It is desirable to make heuristic function considering obstacles, but practically it is very hard to make computationally light heuristic function that meets admissibility condition in every situation. In this reason, most of the A* algorithms use minimum costs without considering obstacles.

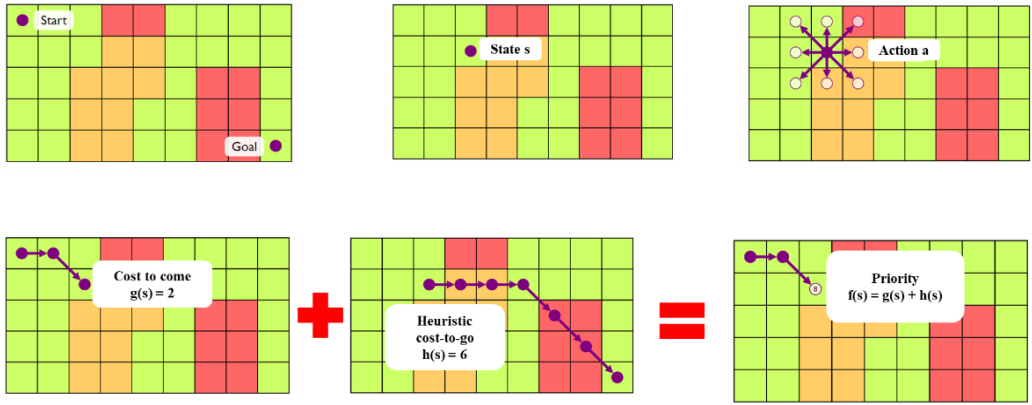

Figure 1.2 A* algorithm explanation image [32]

### 1.2.2. A* footstep planning

Based on A* algorithm, A* footstep planning has similar sequence. Main components (start node, goal node, state, tree, action set, cost to come g(s), heuristic cost to go h(s), priority cost f(s)) are same with A* algorithm. Except some expressions are different. Since humanoid has two feet and walking is done in a left-foot to right-foot or right-foot to left-foot swing, state 's' can be represented as stance foot position and relative angle about x axis of global fixed coordinate system and foot state(left or right). Action set 'A' is a fixed combination of actions 'a', which is represented as delta position and delta angle about previous state. Figure 1.3 describes the feature of action and action set. Also, action set 'A' is mirrored for the opposite stance foot. These actions are expanded as a new search state candidate if it is feasible (not colliding with environment or itself or fits in a steppable region).

*State s = (x, y, $\theta$, foot state)

*Action a = ($\Delta$x, $\Delta$y, $\Delta\theta$)

*Fixed set of footstep actions A= {$a_1, a_2, ..., a_n$}

As you can see in Figure 1.4 footstep planning occurs similarly as an A* algorithm. It expands actions from the start node to the goal node by priority of priority function value of search tree's edge states. Difference is that action set differs by the foot state.

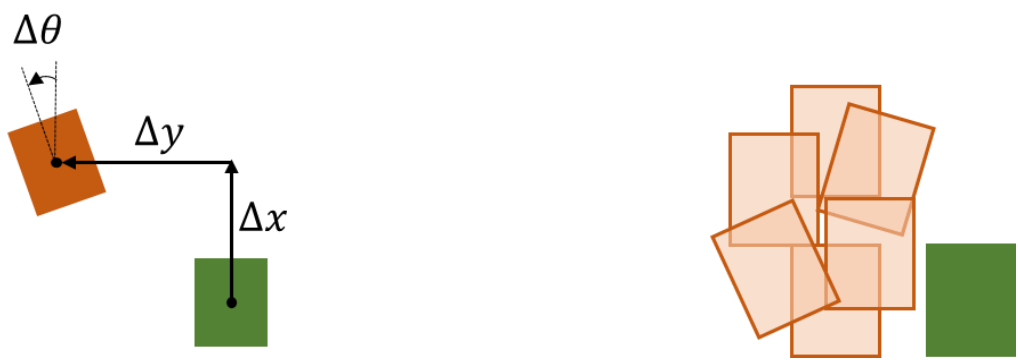

Figure 1.3 Action and action set for footstep planning

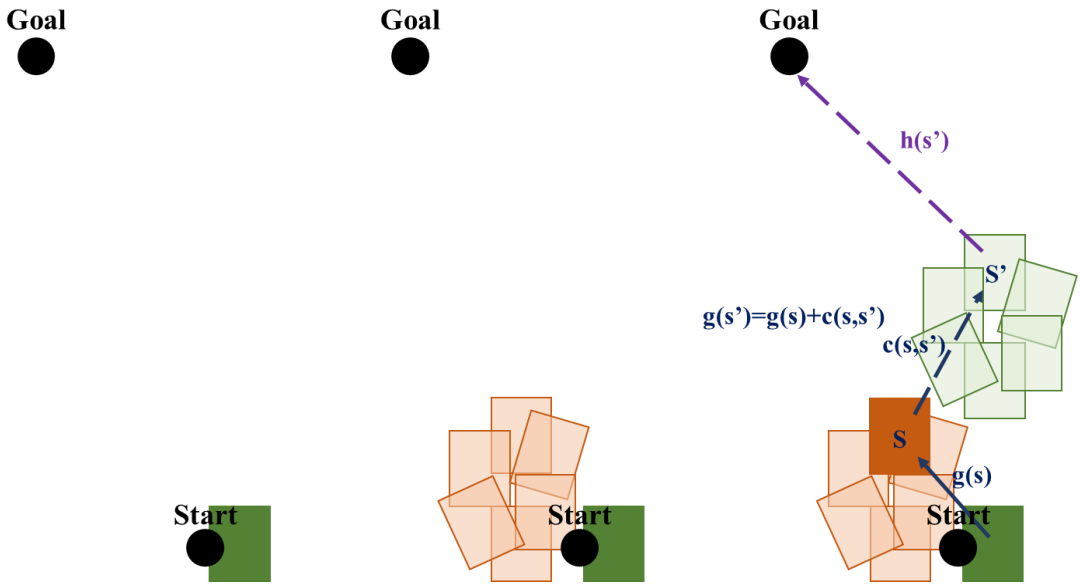

Figure 1.4 A* based footstep planning sequence

Including the basic algorithm of A* footstep planning, it has some key features to concentrate.

- Setting reliable action cost function and heuristic cost function.

  Usually action cost function is defined like Equation (4) as Euclidean distance or additional cost about angle and constant step cost.

$$c(s, s') = \sqrt{\Delta x^2 + \Delta y^2} \quad \text{or} \quad \sqrt{\Delta x^2 + \Delta y^2} + |\Delta\theta| + const \tag{4}$$

Also usually heuristic cost function is defined like Equation (5) as Euclidean distance from state to goal node because of admissibility condition.

$$h(s') = \sqrt{(x_{goal} - x_{s'})^2 + (y_{goal} - y_{s'})^2} \tag{5}$$

It is available to get a possible footstep plan using these cost functions. But since it does not properly express the energy consumption of humanoid, the result may be energy inefficient footstep plans.

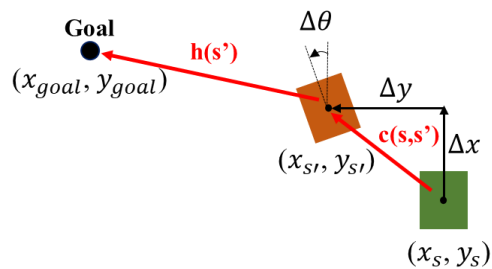

Figure 1.5 Action cost function and heuristic cost function

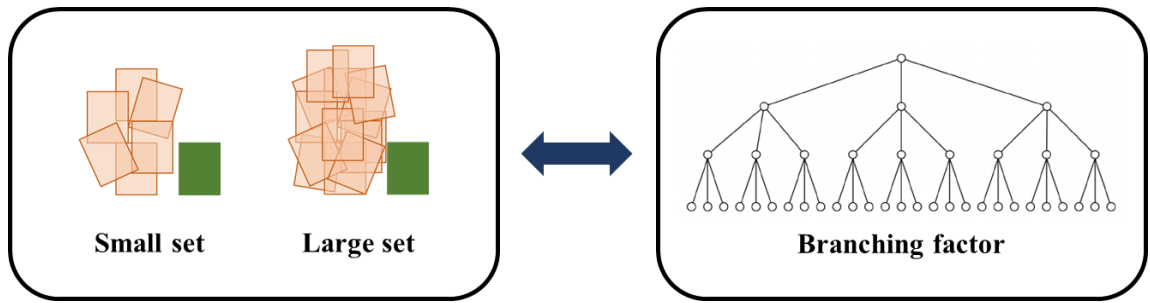

Figure 1.6 Effect of number of fixed actions

- Total number and variety of fixed actions.

Using small number of fixed actions makes branching factor (factor about how much the search tree grows) low but feasibility and coverage of space bad. Contrary, using large number of fixed actions makes feasibility and coverage of space good but branching factor big. In this reason, as your computation power allows, setting the number of fixed action as large as possible is good.

- Feasibility checking methods.

To consider the feasibility of the footstep, it is desired to check with the environment data if a collision occurs or if the ground is too rough to step. Figure 1.7 describes such situations. For more realistic plan, it should consider not only foothold feasibility, but also body feasibility.

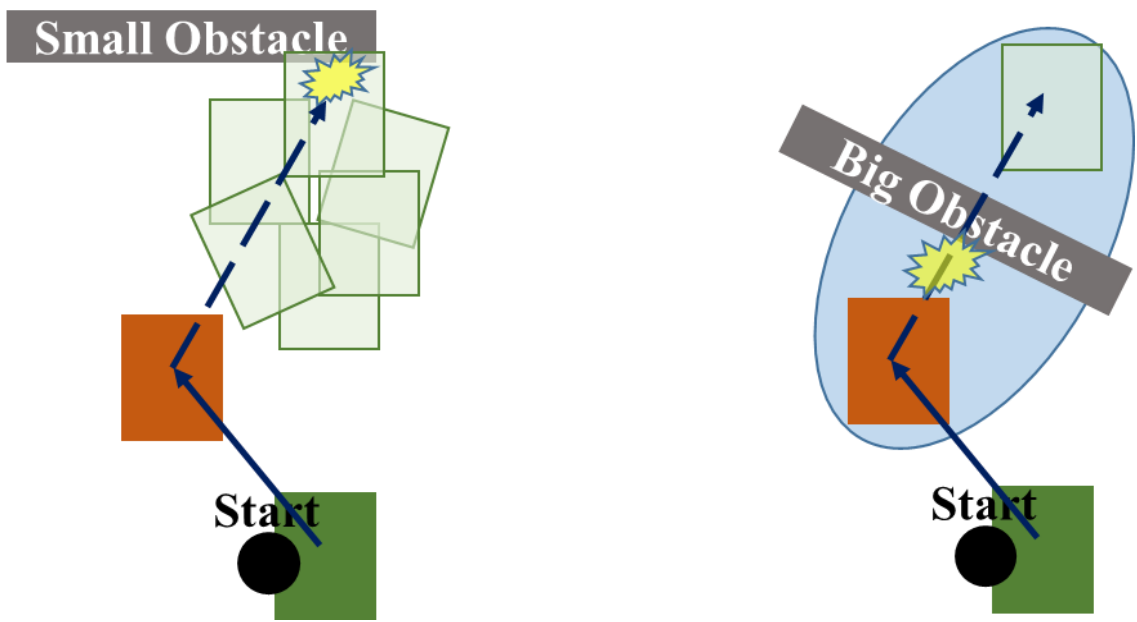

Figure 1.7 Left figure explains the feasibility about small obstacles and rough ground. Right figure explains the feasibility about big obstacles

- Dealing with local minima problem.

It is very hard to define a heuristic function that considers obstacle perfectly. Also, since humanoid is not omnidirectional, it has infinite cases of angles per each position of the map. These two reasons make planning difficult when big obstacle is between robot and the goal. Figure 1.8 explains the situations like it.

Goal

h(s')

Obstacle

Local Minimum

Start

Figure 1.8 Example of local minimum problem

In the point of view of above features, scientists tried to improve each of them.

## 1.3. Related Research

For the cost functions, Huang [14] tried to model the COT (cost of transportation, defined as Equation (6)) with polynomial functions for the action cost function. It brought the idea of function models from the human research papers and validated their model by simulation program. But it used COT as the cost function which made it impossible to model the energy consumption of walking on the spot situations. Additionally it used just Euclidean distance as the heuristic function. It meets the admissibility condition but since the unit is different, it was too small than the action costs which made the search time very long.

$$\text{COT} \triangleq \frac{E_{in}}{m \cdot g \cdot distance} \qquad (6)$$

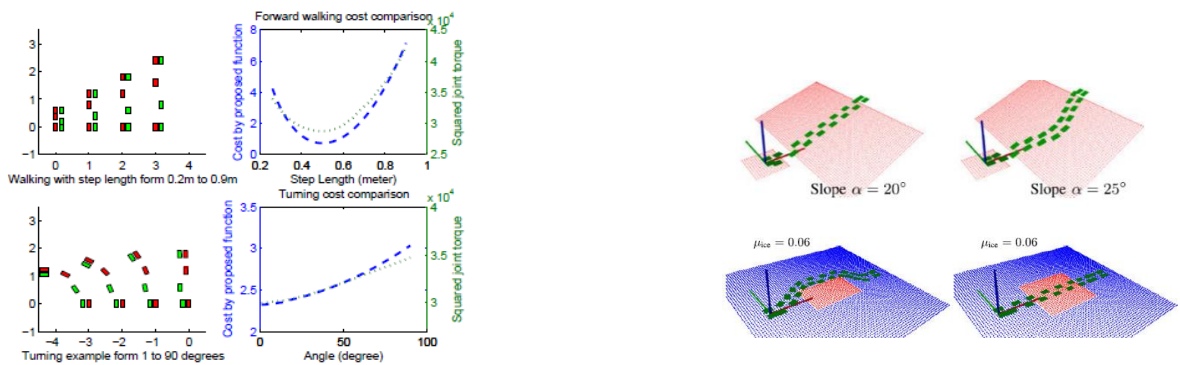

Figure 1.9 Cost function related to energy and its effects. Left figure is Huang [14]'s energy functions validated by dynamic simulations. Right figure is Brandao [16]'s algorithm effects in slope and slippery terrain

Similarly, Brandao[16] let COM energy as action costs which is learned from machine learning with dynamic simulator and heuristic cost as Euclidean distance and weight multiplied with minimal COT. It showed its effect on slippery ground and inclined ground. However, it did not consider the angle difference towards the goal as heuristic function parts.

For the action set number problem, Chestnutt[17] tried a method using 'adaptive action set'. In usual situations, it uses appropriate fixed number of action sets, but when those sets make collision with the environment, adaptively expand more number of action sets. Although it could not solve perfectly the issue of setting the number of action sets, it proposed a better solution than just setting only fixed number of action set. Similarly, Karkowski[18] proposed adaptive action set method, which is almost same in result, but sequentially different. This method works exactly opposite to the Chestnutt's method. For each interval of angles, check if farthest footstep is available, if not, rotate gradually until available, if not, pull the footstep more near and keep on this strategy for each interval. Finally, each interval has only one possible footsteps so total number of actions are small which makes branching factor smaller than big action set.

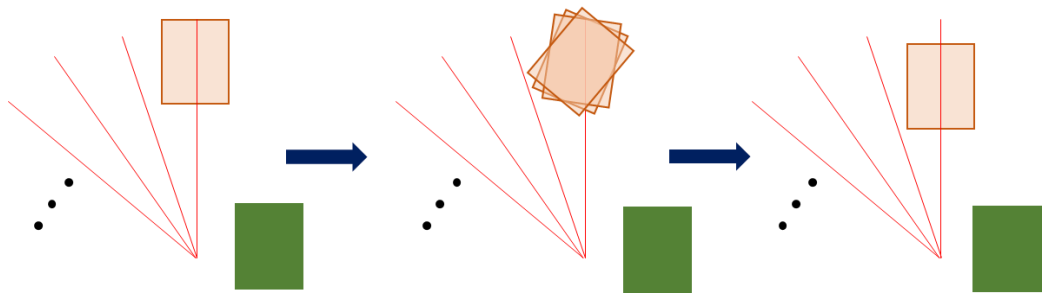

Figure 1.10 Karkowski's Adaptive action set [18]

For the feasibility checking problem, the most common method is called "bounding box method" [22]. It plans a 2d continuous motion of big box that contains the whole robot, and then a sequence of footsteps that can follow the box trajectory is made. However, it cannot consider the very important advantage of humanoid that it can step over or step on an obstacle. Another method is single swept volume method [5]. Based on precomputed body approximations at discrete configurations, they smooth the determined half step combination to get an approximately dynamic movement. Since they approximated the robot movement in high dimensional search space, they had to take into account a random solution using an RRT method instead of A*. These methods made their process faster, but it still needs about 2.5 seconds to make a plan. In addition, they do not use onboard vision sensor. Instead, they use known environment and motion capture device to get localization data.

For the local minimum problem, most of the scientists tried to fusion 2d mobile a* algorithm with A* footstep planning. Chestnutt [19] and Hournung [20] proposed methods to use 2d mobile path as heuristic function of footstep planning. However, it could not consider the ability of stepping over of stepping on an obstacle. LOLA team [21] tried to utilize stepping over or stepping on ability, but it did not consider the energy point of view costs.

## 1.4. Research Objectives

Energy related costs and adaptive action sets are definitely proper research direction for Humanoid A* footstep planning. But there are no full framework that integrates energy related cost functions and adaptive action sets properly, also in the feasibility and local minimum problem points of view, real time application to the real robot is still a challenging problem. Therefore, in this paper, we propose a full framework that integrates energy related cost functions and adaptive actions sets while considering angle difference towards the goal. Furthermore, efficient multilevel feasibility checking method and tricks to reduce local minimum problem is introduced. Finally, integrating framework for 'Mapping' and 'Footstep Planner' and 'Walking Algorithm' will be proposed.

## 1.5. Thesis Overview and Assumptions

In the rest of the thesis, we will introduce and show results in simulation and real robot the full framework of our footstep planner integrated with mapping algorithm and walking controller. Chapter 2 describes the full footstep planner framework, which uses energy related cost functions, adaptive action sets, dual-level feasibility checker and local minima reducer. Chapter 3 describes Integration of mapping, planning, walking. Chapter 4 shows the results on the simulator Gazebo. Chapter 5 shows the results on the real size humanoid 'Gazelle'. Finally, chapter 6 for the conclusion of this paper.

Throughout the thesis, some assumptions are made to focus on the planner itself.

- Self-collision, kinematic constraints, dynamic stability constraints are approximated by step length, width, angle constraints.
- Step time and step height are constant.

# Chapter 2. Footstep Planning with Energy Related Cost Functions and Adaptive Action Set

## 2.1. Energy related cost functions and heuristic functions

### 2.1.1. Action cost function

As Huang proposed [14], Humanoid energy consumption $E_{in}$ can be approximated as a polynomial functions of step length, step width, step angle which is defined as Figure 2.1. With an assumption that it can be separated as Equation (7)

$$E_{in} = E_{length} + E_{width} + E_{angle} \tag{7}$$

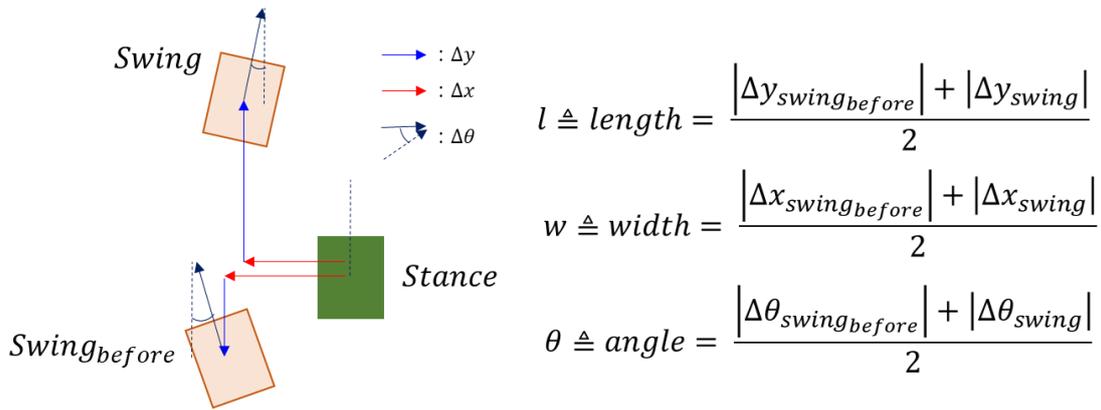

$$l \triangleq length = \frac{|\Delta y_{swing_{before}}| + |\Delta y_{swing}|}{2}$$

$$w \triangleq width = \frac{|\Delta x_{swing_{before}}| + |\Delta x_{swing}|}{2}$$

$$\theta \triangleq angle = \frac{|\Delta \theta_{swing_{before}}| + |\Delta \theta_{swing}|}{2}$$

Figure 2.1 variable definition for energy function

Each Energy functions can now be defined as Equation (8)

$$E_{length} = mg(Al^4 + Bl + C), \quad E_{width} = mg(Dw^2), \quad E_{angle} = mg(F\theta^2) \tag{8}$$

In this equation, A, B, C, D, F are model parameters which is defined by the user. Huang used $E_{length}$, $E_{width}$ equations from human dynamics literatures, but $E_{angle}$ is just defined as above. In order to give more physical meaning to this function, one assumption is made as below.

*To reduce discontinuity in acceleration profile, trajectory is planned with trigonometric function.

While satisfying this assumption, since our system has constant step time, and changing foot yaw angle is dominated by hip yaw joint, energy consumed for changing certain yaw angle in given constant step time is proportional to the square of that certain yaw angle. This is proved in Appendix A.

Furthermore, since we need to consider the view angle of vision sensor, we add an additional penalty costs for sidestep like Equation (9).

$$\begin{aligned} if\ sidestep &\rightarrow\ E_{side} = mg(10*C) \\ else &\rightarrow\ E_{side} = 0 \end{aligned} \quad (9)$$

Now the final action cost function c(s',s) is as Equation (10).

$$c(s, s', s_{bef}) = E_{length} + E_{width} + E_{angle} + E_{side} \quad (10)$$

### 2.1.2. Heuristic cost

To have higher heuristic cost value while meeting the admissibility condition, Brandao proposed [16] to use Euclidean distance multiplied with minimum COT value. COT value is a sort of energy efficiency in the travel distance point of view and defined as Equation (11). Finally, proposed equation for heuristic function is as h(s) in Equation (12).

$$COT = \frac{E_{length} + E_{width} + E_{angle} + E_{side}}{m \cdot g \cdot l} \quad (11)$$

$$\begin{aligned} \alpha &= minimum\ COT\ for\ straight\ walking \\ \Delta d_{goal} &= Euclidean\ distance\ from\ robot\ to\ goal \\ h(s) &= \alpha \cdot m \cdot g \cdot \Delta d_{goal} \end{aligned} \quad (12)$$

Now Euclidean distance has become a cost in order of energy and admissible. However, this heuristic function does not consider the effect of angle difference. Real robot must rotate its body to take an energy efficient step towards the goal like Equation (13). Since our action cost function has a term $E_{angle}$, which is only about angle difference in the assumption of independency, we can form the problem as how much angle difference should be reduced at each step when parameter 'F' is positive number (since it is physical parameter). Some possible solution may be like Figure 2.3. To mathematically prove the optimal reducing method, this problem is formulated and solved with proof as Figure 2.4. Since the method for dividing equally the desired angle is optimal, it is admissible about the given cost of actions while considering angle differences.

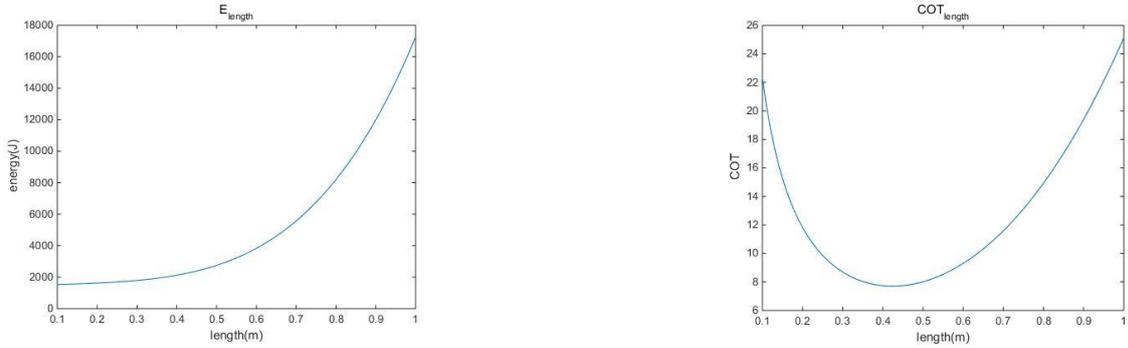

Figure 2.2 Example of $E_{length}$ and $COT_{length}$

$$h(s) = \alpha \cdot m \cdot g \cdot \Delta d_{goal} + func(\Delta \theta_{goal}) \; ; \; E_{angle} = m \cdot g \cdot (F\theta^2) \tag{13}$$

Therefore, the final heuristic function is as Equation (14).

$$h(s) = \alpha \cdot m \cdot g \cdot \Delta d_{goal} + \frac{m \cdot g \cdot F \cdot (\Delta \theta_{goal})^2}{N} \tag{14}$$

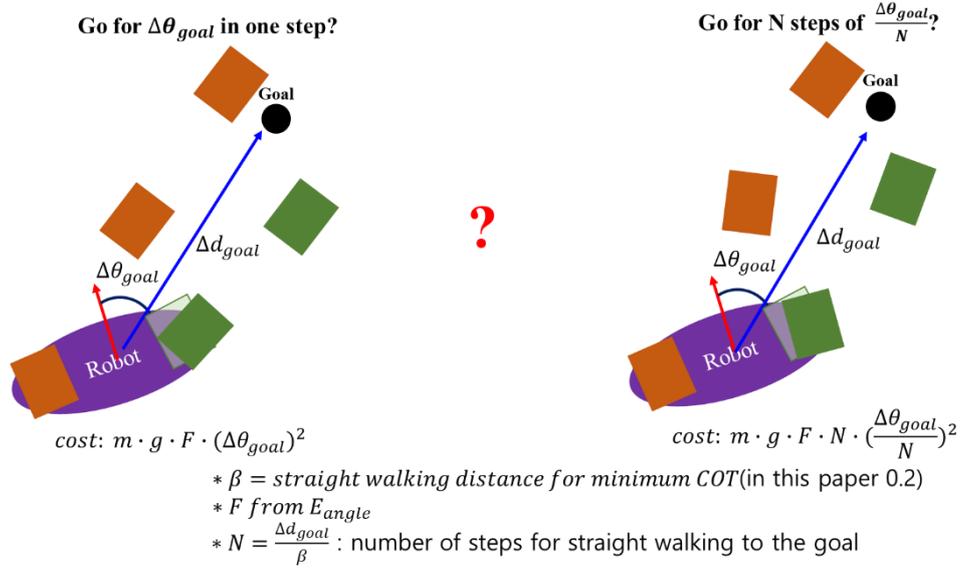

Figure 2.3 Possible solutions for heuristic functions about angle difference

**Problem:**

$$\min_{a_1, a_2, \ldots, a_N} \sum_{n=1}^{N} a_n^2$$

**subject to:**

$$\sum_{n=1}^{N} a_n = \Delta \theta_{goal}$$

**solution:**

$$a_1 = a_2 = \cdots = a_N = \frac{\Delta \theta_{goal}}{N}$$

**proof:**

$$\text{let } \frac{\Delta \theta_{goal}}{N} = c$$

$$\text{then } \sum_{n=1}^{N}(a_n - c) = \Delta \theta_{goal} - \Delta \theta_{goal} = 0 \quad \cdots \text{①}$$

$$\sum_{n=1}^{N} a_n^2 = \sum_{n=1}^{N}(c + a_n - c)^2 = \sum_{n=1}^{N}(c^2 + 2(a_n - c) + (a_n - c)^2)$$

$$= c^2 n + 2 \sum_{n=\frac{1}{N}}^{N}(a_n - c)_n + \sum_{n=1}^{N}(a_n - c)^2$$

$$= c^2 n + \sum_{n=1}^{N}(a_n - c)^2 \cdots from \text{①}$$

$$\therefore a_n = c \quad \text{for minimum} \quad \sum_{n=1}^{N} a_n^2$$

Figure 2.4 Proof of heuristic function considering angle difference

## 2.2. Adaptive action set and selecting method: COT

Adaptive action set is necessary for A* footstep planning as explained in section 1.3. In this thesis we will use this method based on Karkowski's method [8]. Navigation for humanoid essentially needs $(\Delta x_{robot}, \Delta y_{robot}, \Delta \theta_{robot})$ for robots center movement. Furthermore, in our system, vision sensor has certain field of view (FOV). So moving backwards is not desirable. In this reason, we chose fixed action subsets as left figure of Figure 2.5.

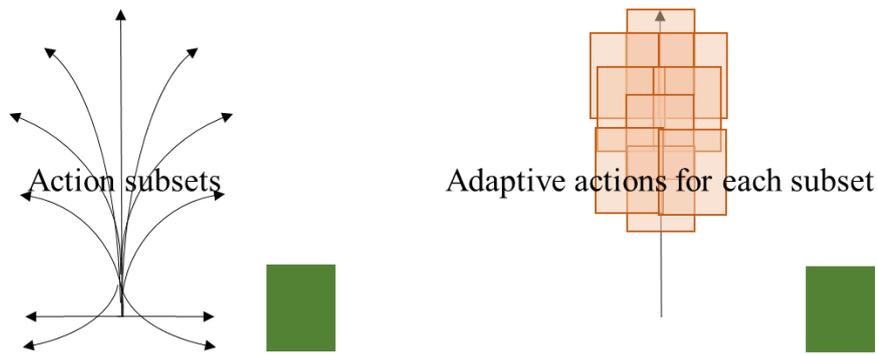

Figure 2.5 Left figure is Action subsets and right figure is example of actions per each subsets

For each subsets, adaptive actions are given like right figure of Figure 2.5. For each subsets we choose one adaptive action which does not collide or is on a safe region. Next thing to consider is which subset to choose when there is multiple action available at certain subset. Previous researchers chose the farthest as possible because their action cost and heuristic cost are about distance values. However, in this thesis those actions are not sure if it is desirable. Therefore, if we assume no obstacles, and enough footsteps are taken, this problem can be defined and solved like Figure 2.6 with proofs. So the minimum COT value action should be selected for each sub action set like Figure 2.7.

*Problem:*
$$\min_{\Delta l}(g(state\ to\ goal)) = \min_{\Delta l}(g(s, s_{goal}))$$

*Assumption:*  1) Straight walking towards the goal without obstacle.
  2) Step number is large enough that constant step size can represent overall step costs.

*solution:*
$$\Delta l = \Delta l\ that\ has\ minimum\ COT$$

*proof:*

**from assumption 1) & 2)** $g(s, s_{goal}) \approx NE_{length}(\Delta l) \quad \leftarrow (N = \frac{euclidean\ distance\ from\ s\ to\ s_{goal}}{\Delta l})$

**let** $d(s, s_{goal}) = euclidean\ distance\ from\ s\ to\ s_{goal}$

**then,** $NE_{length}(\Delta l) = d(s, s_{goal})\frac{E_{length}(\Delta l)}{\Delta l}$

$d(s, s_{goal})\frac{E_{length}(\Delta l)}{\Delta l}$ has minimum value when $\frac{E_{length}(\Delta l)}{\Delta l}$ has minimum value since $d(s, s_{goal})$ is constant as given.

From definition of $COT \triangleq \frac{E_{in}}{m \cdot g \cdot distance}$

$m, g$ is given constant $\quad \rightarrow$ minimum COT is equivalent to minimum $\frac{E_{in}}{distance}$

$\therefore \min_{\Delta l}(g(state\ to\ goal)) \quad \rightarrow \quad$ **when** $\Delta l = \Delta l_{minimum\ COT}$

Figure 2.6 Proof that adaptive action selection method should be minimum COT value

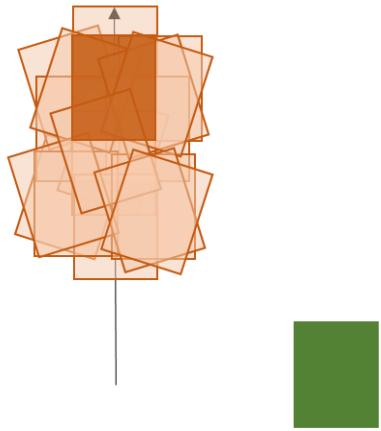

Lowest COT action chosen

Figure 2.7 Final adaptive action selection method: minimum COT

Additionally, when state is near the goal node, selecting minimum COT value does not always make the robot reach the goal position even if there is no obstacle. Figure 2.8(a) is one of the examples of those situations. To overcome this problem, when robot's center gets near to the goal for a certain length, stop using adaptive sets but use full sets like Figure 2.8(b)

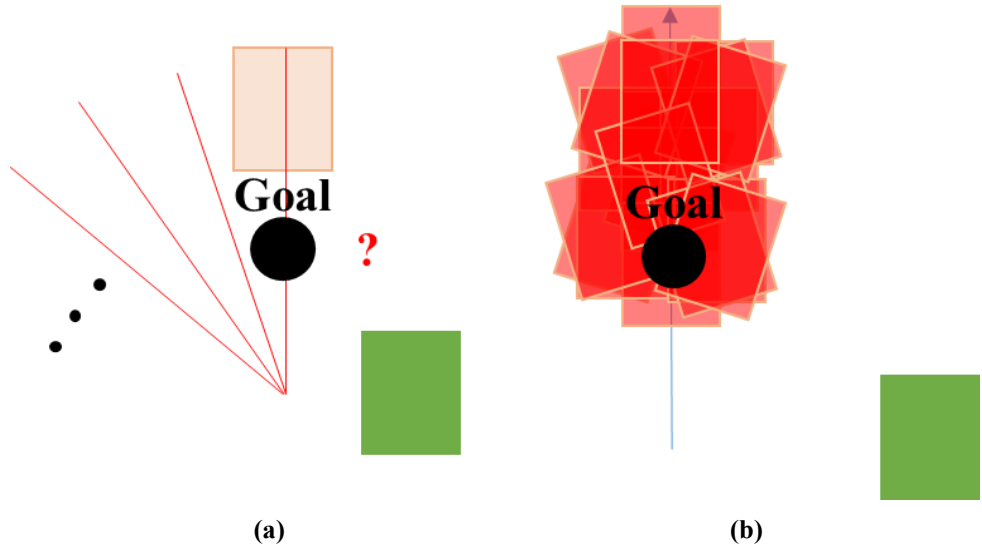

(a)　　　　　　　　　　　　　(b)

Figure 2.8 Near goal adaptive action set problem and solution

## 2.3. Dual-level efficient feasibility check

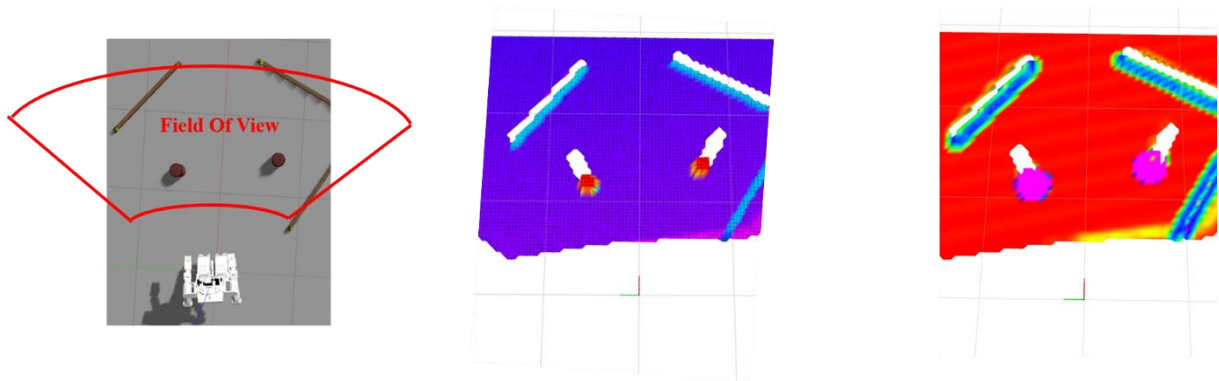

Figure 2.9 Elevation Map and Traversability Map example

Feasibility avoidance is one of the most important feature in footstep planning. In this thesis, to deal with the computational problem, feasibility is checked in a grid map space by a dual level height check mechanism. Before heading on the specific algorithms, grid map values used for checking the feasibility will be introduced. Overall grid maps are as Figure 2.9. It is originated from grid-map package and Elevation-mapping package (open source ROS Package developed at ETH) [28]. Specific usage of data structures of these packages and how the values are calculated will be introduced at chapter 3. With these maps, feasibility checking algorithm is done by first checking the stepping region of the action for its feasibility. If any grid cell inside the region of the footstep has bad feasibility or has higher height than 5cm, it is considered as unavailable. If it is available for the footholds, it now checks about its body collision. For computational reason, body collision of the swinging motion will be approximated as a sum of ellipses. Finally, two ellipse region is checked for the body collision. One between the stance foot and new swing foot. Second between the before swing foot and new swing foot. Since the real robot

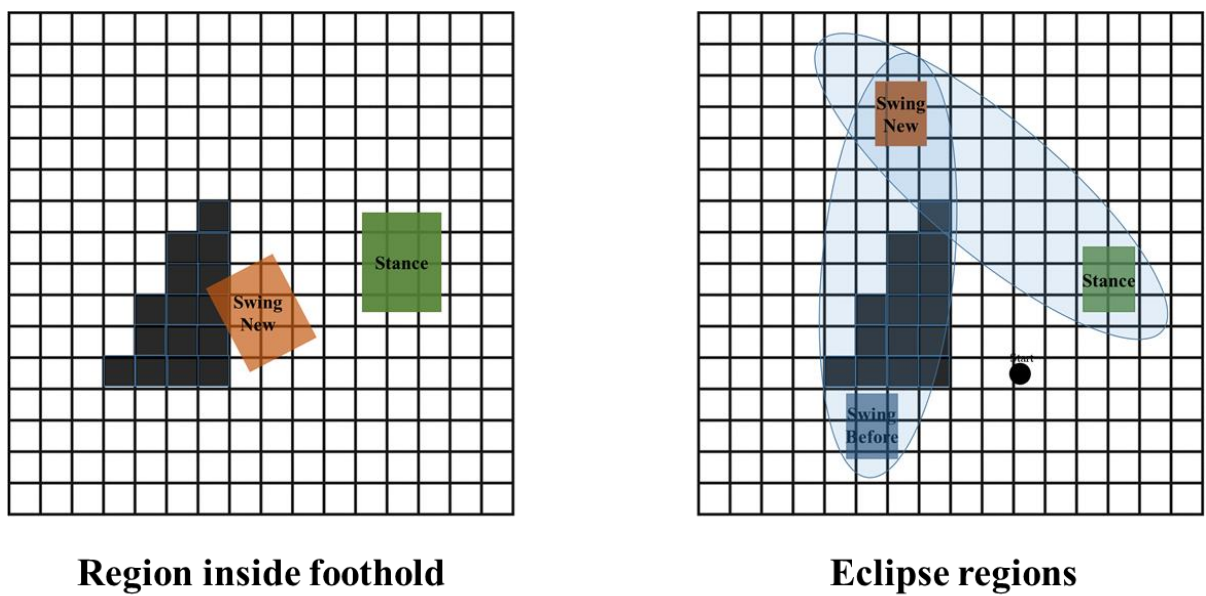

**Region inside foothold**                **Eclipse regions**

Figure 2.10 Dual Level Feasibility Checking Methods. Left figure explains the checking region for low obstacles and rough surfaces. Right figure explains the checking region for higher obstacles

sways in breadthwise direction, additional interval is considered in the longer axis parameter of the first ellipse. Assuming that both ellipse has same shorter axis parameter, and since the ellipse between before swing foot and stance foot will be checked at the before state, checking these two ellipse regions for high height data is enough for checking the body collision properly.

## 2.4. Penalizing Heuristic Costs to Reduce Local Minimal problem.

To reduce the local minima problem near the big obstacles, we use some triggering penalty costs for heuristic cost function. The algorithm works when there is an obstacle in the direction of the new action. If obstacle is closer than a certain distance is the direction, it gives penalty to the heuristic function so those footholds will be searched later than the other states. Figure 2.11(a) describes those situations. Additionally the only rotating and staying still at the same position is less penalized because it is an action to avoid the obstacle. Lastly to avoid the stepping backwards situation (rotating and staying still action has very little backward moving property), if the before step was rotating and staying still action with the opposite direction, it is not less penalized, but largely penalized. Figure 2.11(b) describes those situations.

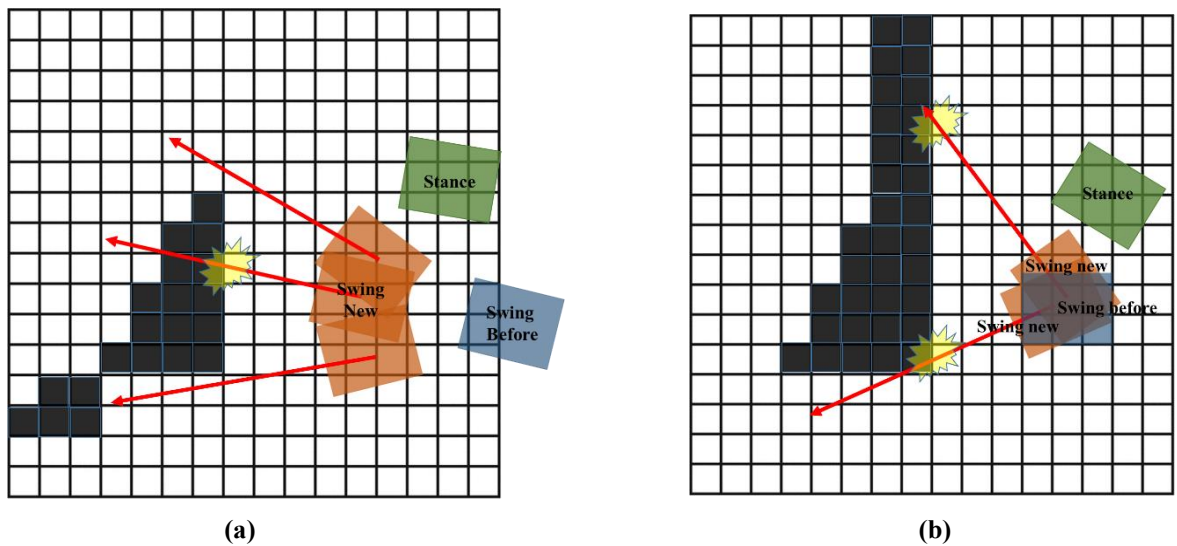

(a)  (b)

Figure 2.11 (a) describes the checking for obstacles in a certain distance in a direction of the new foothold. (b) describes the high penalty for going backwards

Specifically, it only considers the obstacle until the colliding part through the foot direction because of the computation problem. Furthermore, it considers the body collision at that point, which is 0.5m interval for our system. Its specific figure is as Figure 2.13. Heuristic function is penalized with geometrical calculations like Figure 2.12 and final function is as Equation (15)

Figure 2.13 Specific figure of the penalized heuristic function.

Figure 2.12 Geometrical description of penalized heuristic function.

$$h_{penalized}(s) = \alpha \cdot m \cdot g \cdot (d_2 + d_3) + \frac{m \cdot g \cdot F \cdot (\theta_2)^2}{N_2} + \frac{m \cdot g \cdot F \cdot (\theta_3)^2}{N_3} \tag{15}$$

Finally, our footstep planner is integrated in a single framework. It searches the optimal footstep plan within its adaptive action set to minimize human inspired energy functions, while avoiding unsteppable situations and reducing local minimum problems. Overall pseudo code is as Figure 2.14.

```
s_start ← set_start(position_start, orientation_start)
s_goal ← set_goal(position_goal)
Environment ← map_from_data(cameradatas)
openlist ← openlist.pushback(s_start)
while
    Q ← openlist.pop_minimum_f()
    if distance(Q, s_goal) < a then
        closedlist ← closedlist.pushback(Q)
        output_plan ← track_closedlist(Q)
        break
    end
    else      then
        closedlist ← closedlist.pushback(Q)
        actionlist ← init_actionlist(actionlist)
        for i ; i < number_of_subset ; i++
            possible_adaptivesets ← collisioncheck(Q, adaptivesets)
            actionlist ← actionlist.pushback(minCOT(possible_adaptivesets))
        end
        openlist ← openlist.pushback(actionlist)
    end
end
```

Figure 2.14 Pseudo code of footstep planning

# Chapter 3. Integrating Framework: Mapping, Planning, Walking

Real time Humanoid Navigation with footstep planning is done in three stages. It first makes a global map about a global world coordinate, which includes environment data. Next, it makes a footstep plan with the environment data and its certain cost functions. Finally, with given footsteps, humanoid makes an appropriate locomotion. To make real time feedback about environment, footstep plan should be updated at least each step. Step time of our algorithm is fixed as one seconds, so at least, map update and footstep plan update should be done in one seconds.

Resulting framework and data flows are as Figure 3.1. PODO is our Humanoid's original software platform [24]. It stores core data needed for humanoids in shared memory and main walking algorithms are run inside this platform and is run in a mini PC. Other softwares like mapping and footstep planning is run in a ROS platform, which is very large and frequently used open platform [25]. It is run in a different computer with GPU integrated. Robot's sensor data except camera (encoders and IMU and FT sensors) are directly read in PODO system. These data are used to make odometry data, which is estimation of robot's pelvis pose and orientation. Later, odometry data and encoder data are transferred to ROS platform and are used to generate TF data, which is full coordinates of each links of the robot about global world coordinate Camera data are directly read in ROS system and are used to generate point-cloud data. Mapping and filtering packages integrates these data to generate map data. Now, footstep planner utilized the map, odometry, TF data to generate footsteps and transfers this to the PODO system's walking controller. Finally, walking controller generates sufficient motor references to follow the footsteps.

Succeeding contents are detailed explanation of 'mapping and filtering packages', 'Synchronization of footstep planner and walking algorithm' and 'Data transfer methods and data structures'.

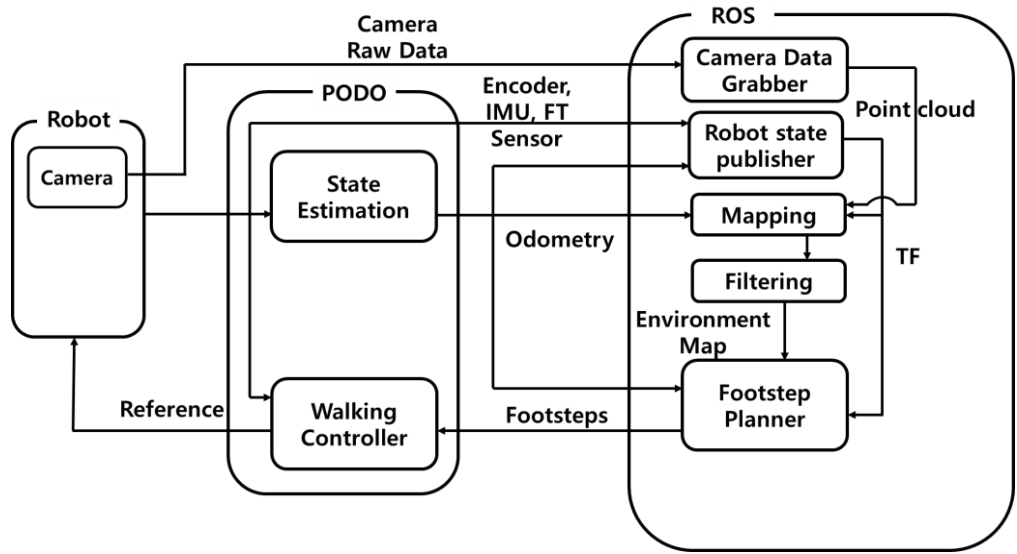

Figure 3.1 Data flow of Real time humanoid navigation by footstep planning

## 3.1. Elevation Mapping and Filtering

### 3.1.1. Elevation Mapping [28]

Elevation mapping is an open-source ROS package, which is a robot centered height mapping algorithm. It needs odometry value and point-cloud data from the camera sensor with TF tree made by encoder values and odometry. It is based on a probability based filter algorithms. Therefore, camera noise and odometry uncertainties are considered when mapping is done. In addition, it is a grid based mapping, so the access to the map data is structured, which accelerates the iterating time of map data structure. Since height-map can only consider 2.5D environments, in this thesis we will only consider those situations. (Example: not considering multi heights in one cell.)

Elevation mapping has various parameters to consider, sensor noise model parameters, grid resolution, grid map size, etc… Each parameters are selected as Table 3.1 and resulting map is like Figure 2.9

Table 3.1 Map parameters

| Mapping parameter | Value |
|---|---|
| Map length in x direction | 3.0m |
| Map length in y direction | 3.0m |
| Resolution | 0.05m |
| Minimum Variance | 0.1 |
| Maximum Variance | 10 |
| Mahalanobis Distance Threshold | 2.5 |
| Multi Height Noise | 0.000002 |

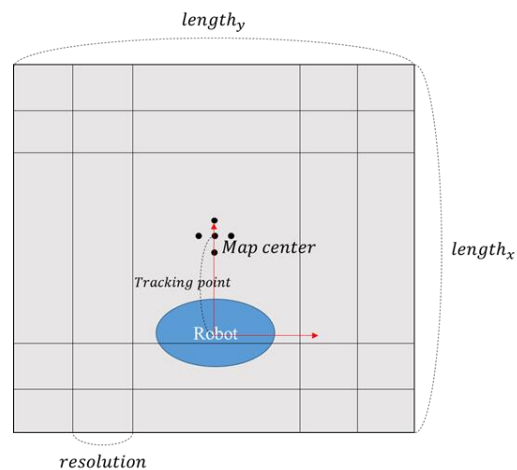

Figure 3.2 Elevation Map Parameters

## 3.1.2. Filtering [29]

Resulting map still has some noise, and mapping is not always perfect. Furthermore, as written as Chapter 2, footstep planner needs not only height values, but also feasibility values of environments to check the footstep feasibility. In this reason, certain filtering is done to the elevation map like average filtering, normal filtering, slope filtering, roughness filtering and feasibility filtering. Each filtering result is as Table 3.2. Finally smoothed elevation map and traversability map values are used for feasibility check in footstep planning.

Table 3.2 Map Filtering Examples

| | | |
|---|---|---|
| Original Elevation Map | | 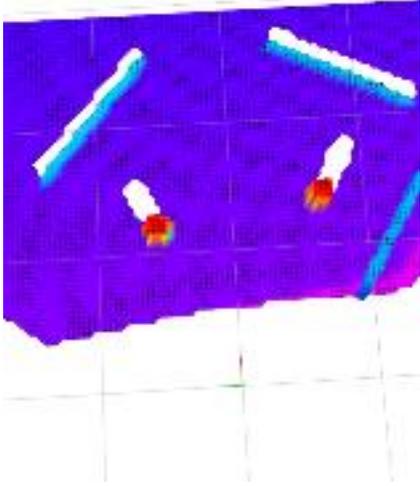 |
| Average Filtered Elevation Map | Radius : 0.1m | 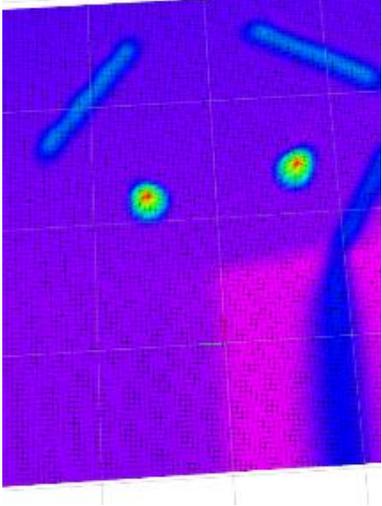 |

| Normal Vector Map | Radius : 0.1m | 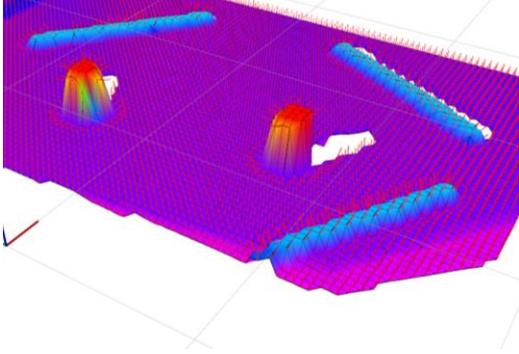 |
|---|---|---|
| Slope Map | acos(Normal_z) | 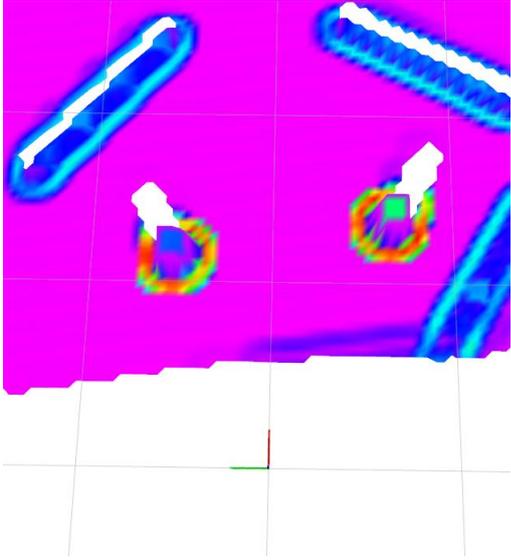 |
| Roughness Map | (Original Elevation Map) – (Average Filtered Elevation Map) | 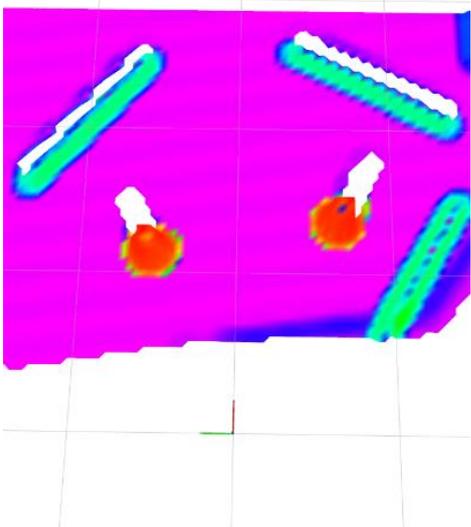 |
| Traversability Map | 0.5* (1.0-(slope/0.6)) + 0.5* (1.0-(roughness/0.1)) *lower threshold: 0.0 *upper threshold: 1.0 | 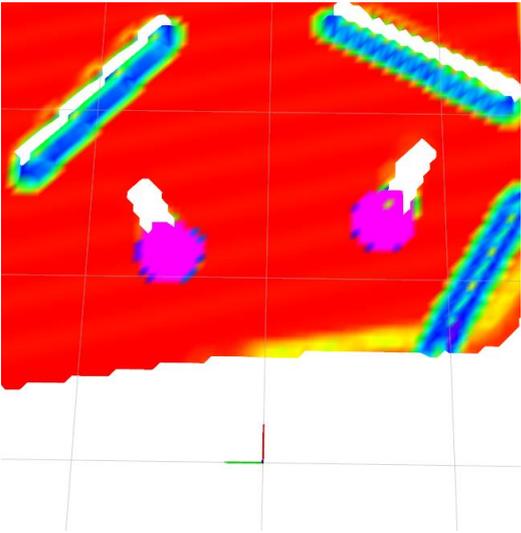 |

## 3.2. Synchronization of Footstep Planner and Walking Controller

Our humanoid robot uses walking algorithms based on Preview Control [35]. This algorithm plans the reference motions considering future footsteps, in this reason it needs most recent continuous three steps as input values in our walking algorithm like Figure 3.3. In addition, to avoid the situations like when robot is on a right foot swing state, but footstep plan is in left foot swing state, we used some synchronizing step numbers to synchronize between robot and footstep planner. These step numbers are counted for each steps of real robot like Figure 3.4 and given to the footstep planner. After that, footstep planner returns the plan and that step number. Therefore, if the both step number is not the same, it means the planner could not give plan before the footstep has changed. In this situation, we give step on the spot steps for safety. Finally, for situations as if goal node is reachable less than three steps, we give step on the spot steps for the rest of the steps. In conclusion, our walking algorithm reads the most recent plan's three steps, which has synchronized step number.

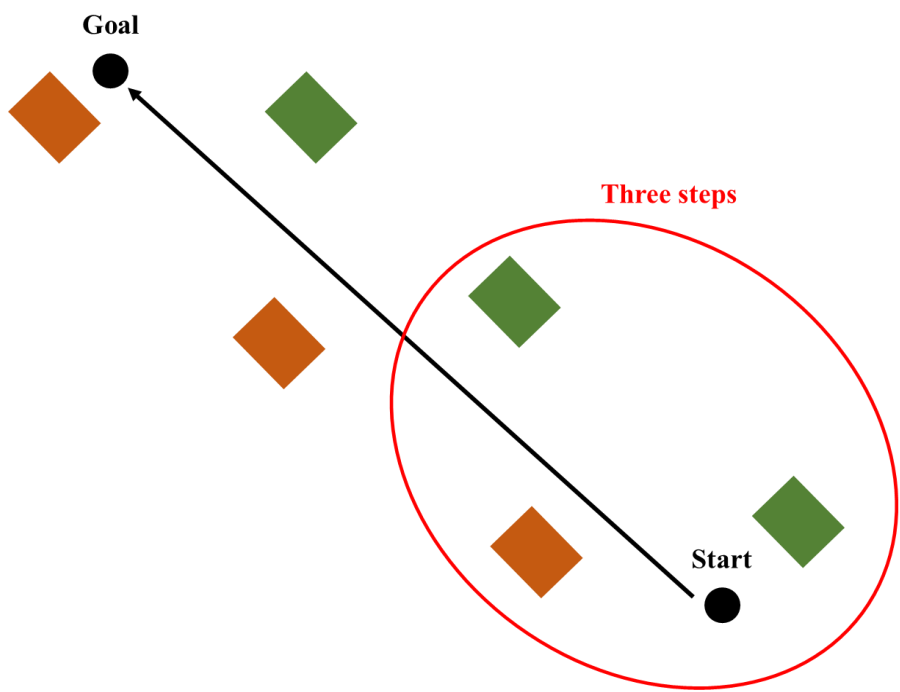

Figure 3.3 Three steps from footstep planner to walking controller needed for preview control

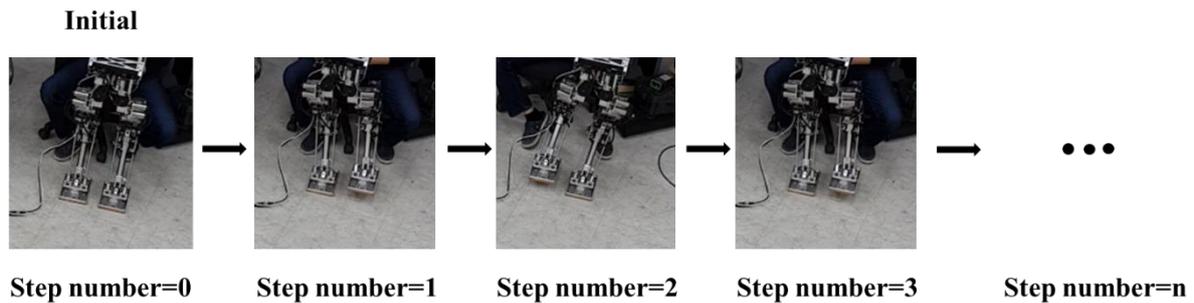

Figure 3.4 Step number counting

## 3.3. Data transfer method: Inside ROS, ROS and PODO

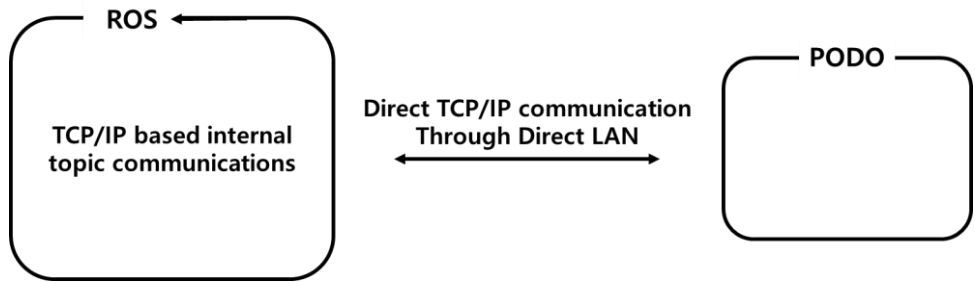

Figure 3.5 Data transfer methods

Data for the whole frameworks are based on TCP/IP communication. ROS uses TCP/IP based communication named topic. In addition, data to be transferred between ROS and PODO are transferred by direct TCP/IP communication through direct LAN cable connection.

### 3.3.1. Inside ROS

ROS has some essential components, Master, Node, Package, Message, Topic, Publisher, and Subscriber. It has more components, but above are enough for essential parts in this thesis. 'Master' is a kind of main server to connect node to node and communication with messages. 'Node' is a minimum unit processor in ROS. When node is activated, it sends the name of node, topic, message structure form, URL address, port to the master. Using these information, each nodes communicates each other with topics. 'Message' is a data structure transferred between nodes. It includes data variables like 'int' and 'double'. 'Topic' is a message with a nickname. Publisher node gives the master the information of the message as a topic, and subscriber node gets the information of the publisher node from the topic registered in the master. 'Publish' means sending the message form data, which is a content of a topic. 'Publisher' node sends the information of itself including the topic information to the master to publish. Similarly, 'subscribe' means receiving the message form data from a publisher. 'Subscriber' node also sends the information of itself including the topic information to subscribe to the master. Using this data, it receives the data of the publisher node from the master and directly connects to the publisher node to receive the desired message data.

In our framework, various topics are published and subscribed like Table 3.3

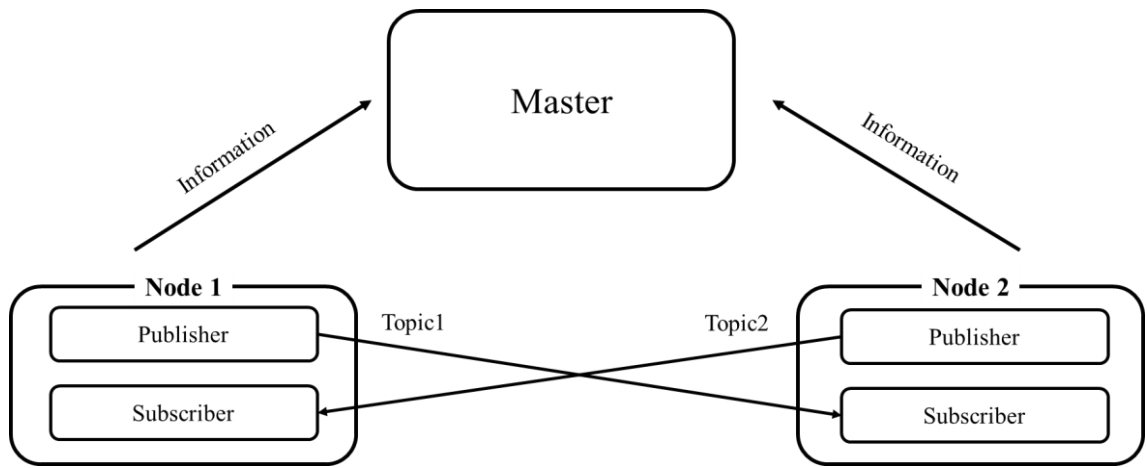

Figure 3.6 ROS data transfer between nodes

Table 3.3 Important Topics

| Topic Name | Update Frequency |
| --- | --- |
| Map data | 2~3 |
| TF | 500 |
| Odometry | 500 |
| Camera data | 30 |
| Footsteps | ~1 |

### 3.3.2. ROS and PODO: TCP/IP

To transfer data to the non-ROS systems like PODO we use direct TCP/IP communication. We made a server at the PODO system and ROS node named 'PODO_Connector' as the client. This node receives certain data from the PODO and sends certain data to the PODO. Simultaneously it publishes and subscribes those data from other ROS nodes. Figure 3.7 explains this data transfer. PODO_Connector receives and sends the data like Table 3.4

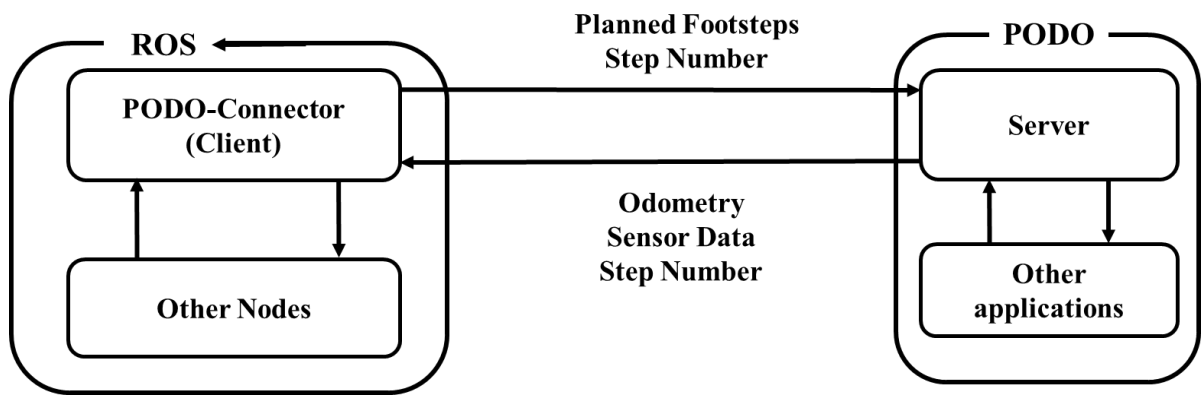

Figure 3.7 Data transfer by TCP/IP. Server is made in PODO system

Table 3.4 Transferred data between ROS and PODO

| Data | Transfer frequency |
|---|---|
| Planned Footsteps | ~1 |
| Step number from planner | ~1 |
| Odometry | 500 |
| Sensor data | 500 |
| Step number from robot | 500 |

Integrating the whole features, the whole real time navigation framework is run as Figure 3.8.

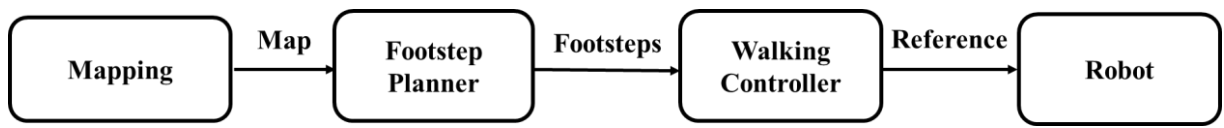

Figure 3.8 Whole Humanoid navigation framework

# Chapter 4. Simulation Results

To validate the frameworks, we first integrated the frameworks with a dynamic simulator. The only difference with the original framework is that the robot part of Figure 3.1 is exchanged to the simulated robot.

## 4.1. Gazebo

Gazebo is a dynamics simulator, which is very commonly used with ROS systems, one of our humanoid DRC-HUBO+ is well simulated in this simulator, so we applied the whole framework with this simulator. DRC-HUBO+ is a humanoid developed for 2015 DARPA Challenge Finals. Its Specification is as Figure 4.1. The robot is simulated in the Gazebo simulator world with KINECT sensor as the RGBD sensor. Its action cost function variables are defined as Table 4.1 and kinematic constraints are set as Table 4.2. Finally, Figure 4.2 is an example of simulation environment and simulated robot.

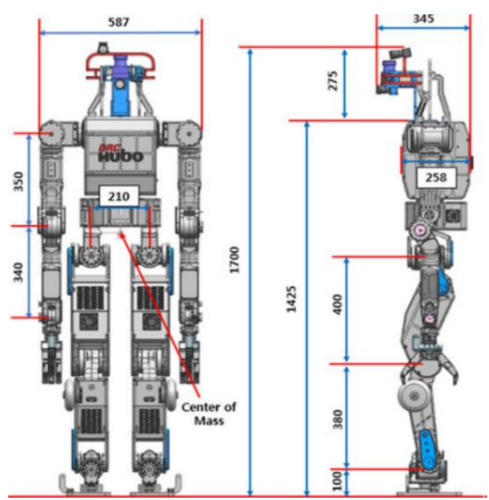

Figure 4.1 DRC-HUBO+ specification [36]

State estimation, which is an approximation of robot's position and orientation about global axis, is done by integrating visual odometry and IMU data.

Table 4.1 Action Cost Function variables

| Energy Cost Function Variable | Value |
|---|---|
| A | 44.0 |
| B | 0.2112 |
| C | 4.0 |
| D | 0.2 |
| E | 0.23 |
| F | 0.4 |

Table 4.2 Kinematic constraints for simulated robot

| Kinematic Constraints | Value |
|---|---|
| Maximum step length | 0.40 m |
| Optimal step length | 0.30 m |
| Maximum step width | 0.35 m |
| Minimum step width | 0.18 m |
| Maximum step yaw | 15 deg |

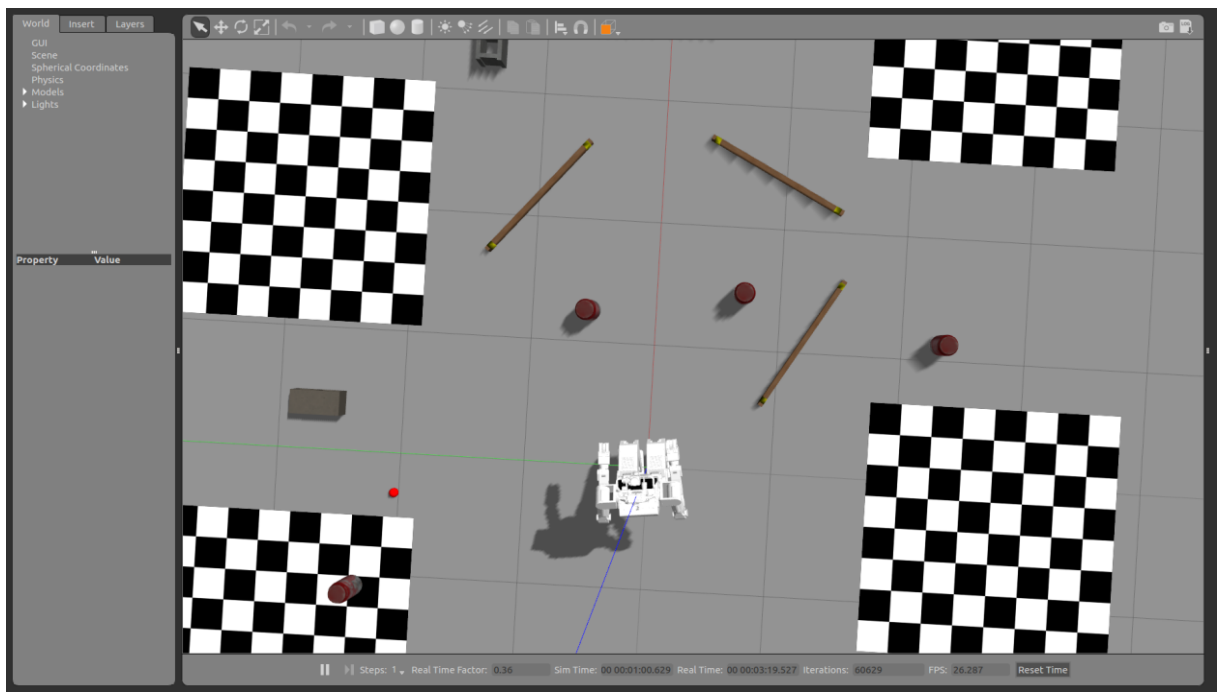

Figure 4.2 Example of simulation environment with simulated robot

## 4.2. Results

1) Verification and effect on cost of choosing least COT action as adaptive actions set selecting method.

Selecting the least COT action for the adaptive action set is proposed as a sufficient way to choose action. It is verified by comparing with choosing farthest action. To satisfy the assumption made in Figure 2.6, goal is selected to have enough distance from start position. The result is as Figure 4.3. As expected, the least COT selection method has less cost. This means least COT method is more sufficient way to choose an action from adaptive action sets for human inspired energetic costs.

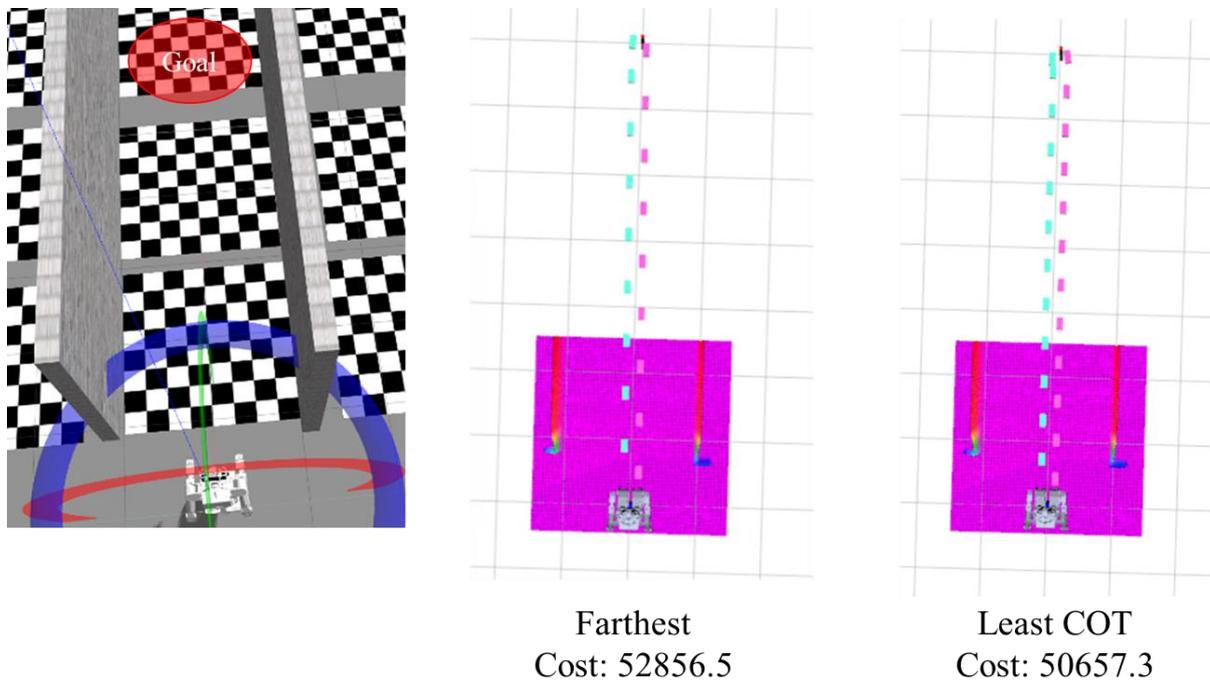

Farthest
Cost: 52856.5

Least COT
Cost: 50657.3

Figure 4.3 Verification of adaptive action set selecting method: least COT

To check if the result is similar in difficult tasks, another result is shown as Figure 4.4. Also, as expected, it shows similar result. Least COT selecting method gives less cost than selecting farthest action.

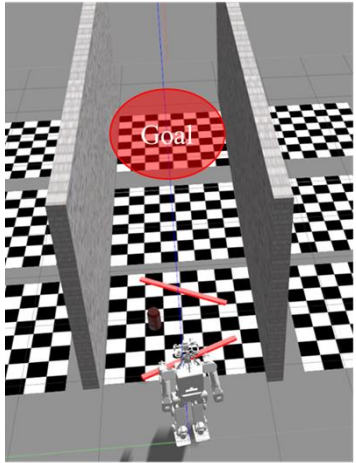 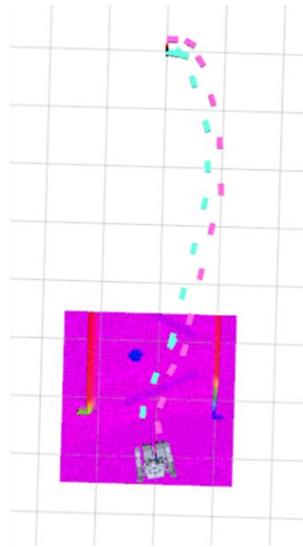 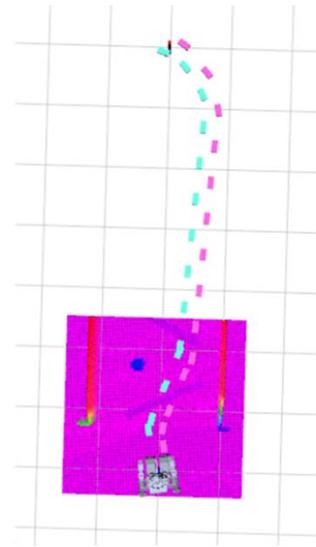

Farthest  
Cost: 56538.4

Least COT  
Cost: 53864.6

Figure 4.4 Verification of adaptive action set selecting method in more complex tasks: least COT

2) Verification and effect on iteration number of heuristic function considering angle difference.

To verify the effect of heuristic function considering angle difference, a task without obstacle and with turning is given. Situation is as Figure 4.5. One of its result is as Figure 4.6. It has almost same final cost with less iteration. Furthermore, various tasks which is similar to the one like Figure 4.5 is done and its average result is as Figure 4.7. As expected, almost same optimal solution is planned with 19.7% less iteration.

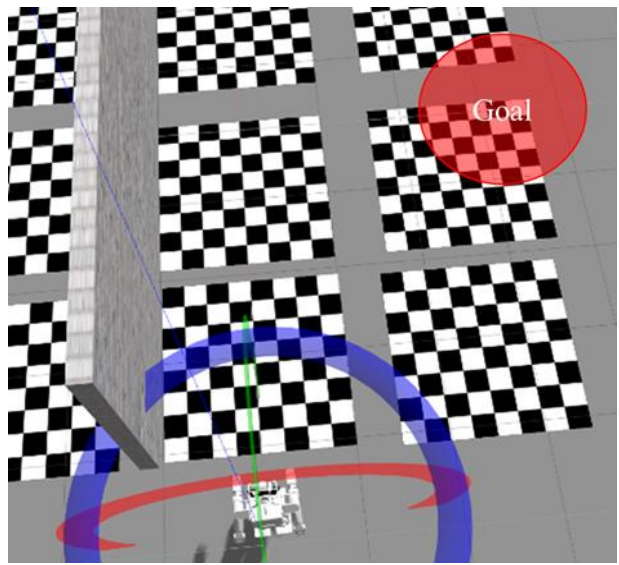

Figure 4.5 Turning in non-obstacle situation task

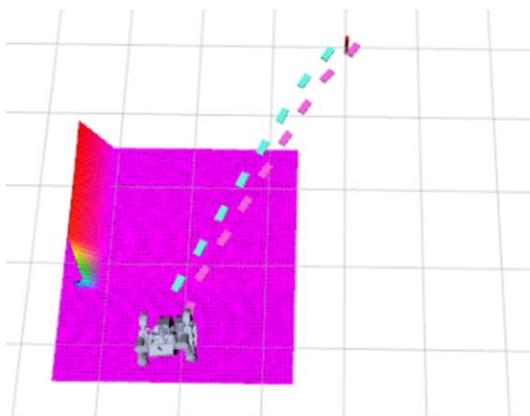
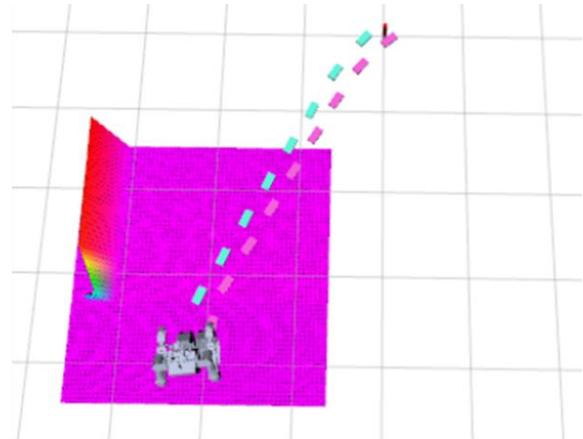

**Heuristic with No Angle**
Final Cost: 40709.9
Iteration: 318

**Heuristic with Angle**
Final Cost: 40785.4
Iteration: 289

Figure 4.6 One example of results for turning

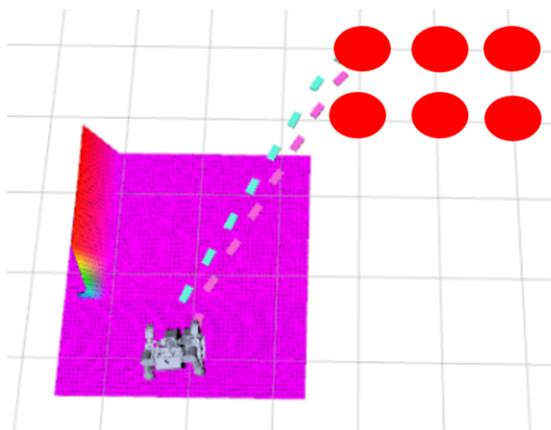

| Average cost difference | Average iteration difference |
|---|---|
| 0.41% | 19.70% less |

Figure 4.7 Various turning situations and its average results

3) Verification and effect on local minimum problems of penalizing heuristic function by geometrical way.

Simplest local minimum problem, which is when big obstacle exists between robot and the goal, is given. The result without and with penalty on heuristic function is as Figure 4.8. As expected, plan without penalty on heuristic function could not solve the local minimum problem within maximum iteration number 2000. However, with the penalty, it was able to make a plan in 617 iterations.

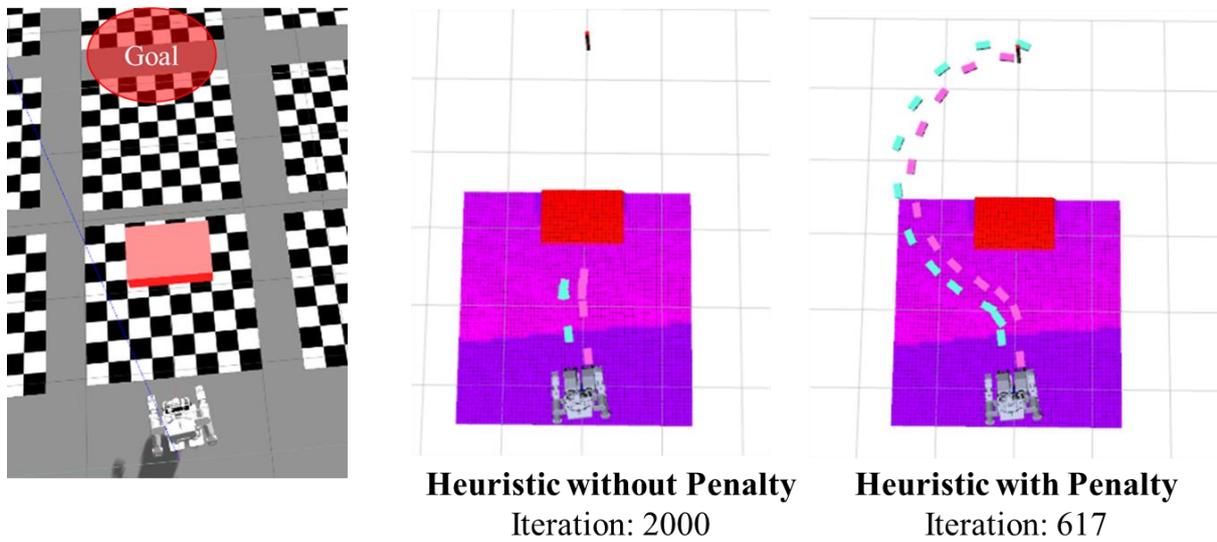

Figure 4.8 Local minimum problem with and without penalty on heuristic function

4) Complex task result

Summarizing all parts, made a full navigation situation like Figure 4.9. It goes through these complex environments real time.

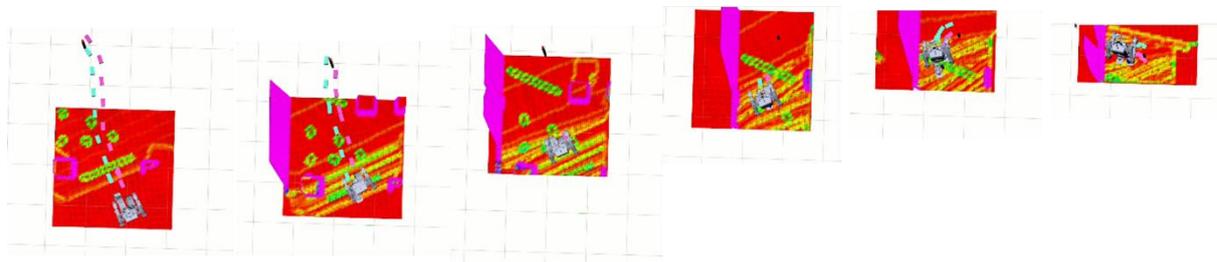

Figure 4.9 Full navigation simulation with mapping and footstep planning and walking

Furthermore, to check if the full framework is possible in the real robot, experiments for real robot navigation is done at Chapter 5.

# Chapter 5. Real Robot Experiment

In this chapter, the results for real robot experiment is explained.

## 5.1. Experimental Setup

### 5.1.1. KINECT V2 [34]

For real robot, KINECT V2 is selected for the RGBD camera sensor. It is first made at 2014 and it has great advantage for its high accuracy in indoor situations. (Where there is no sunlight.) Since our experiment will be done in indoor environments, it is a very suitable sensor for us. It has three kinds of resolution setting for depth data, hd, qhd, sd. Considering the accuracy and the computation time, qhd resolution is selected for the setting. In addition, calibration is needed to synchronize between the infrared-based depth and RGB based depth data, where KINECT V2 uses both data to produce final depth data. Calibration is done by minimizing the sum of difference of checkerboard data. It is done as below sequence. Finally resulting maps with the KINECT V2 sensor is as Figure 5.2.

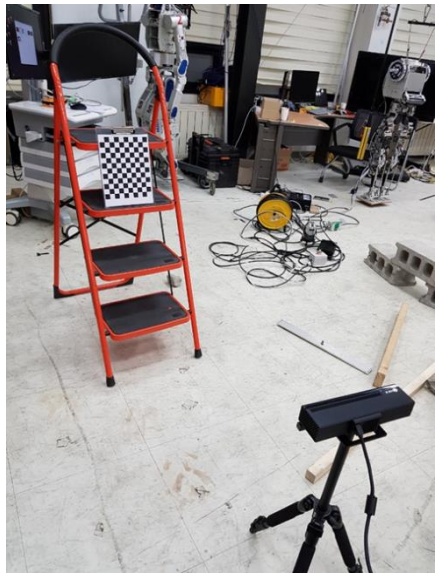

Figure 5.1 Calibrating KINECT V2 with checkerboard.

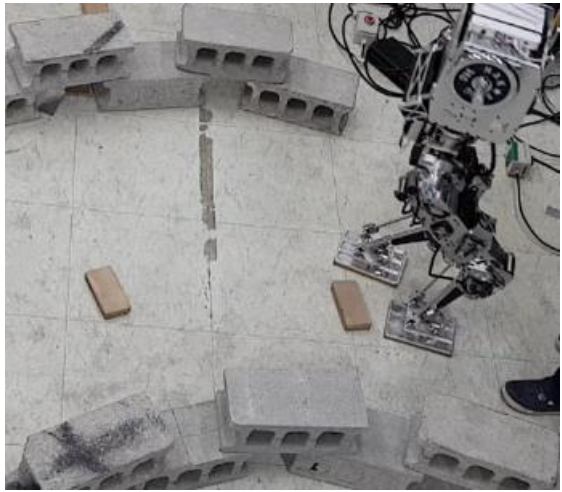
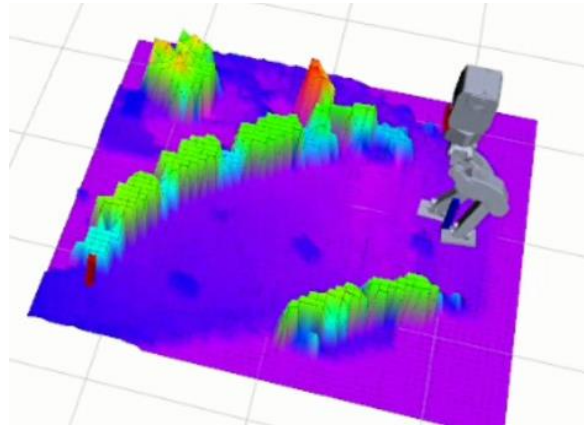

Figure 5.2 Example map from Kinect V2 [34] and elevation mapping [28]

### 5.1.2. Humanoid Platform, Gazelle

'Gazelle' is a humanoid robot based on electronic actuators. It is specialized for bipedal locomotion, so it does not have an arm. Its main feature is that actuators for the ankle is gathered in the upper part to make the inertia small. Its specifications are as Figure 5.4. Additionally, considering the outer case of the robot, the camera mount is added like Figure 5.3. Its kinematic constraints are set as Table 5.1.

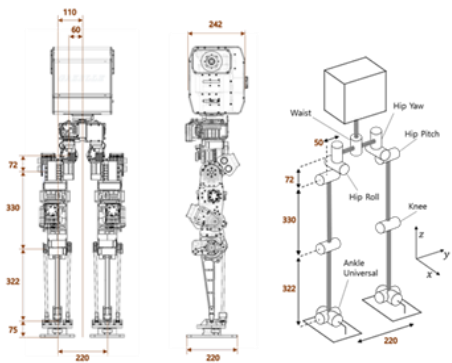

| TABLE I | | |
|---|---|---|
| SPECIFICATION OF GAZELLE | | |
| Height | | 130 cm |
| Weight | Inc. battery | 37 kg |
| | Exc. battery | 33 kg |
| DOF | Total | 13 |
| | Leg | 2 legs × 6 DOF |
| | Waist | 1 DOF |
| Sensors | | IMU (gyro and accelerometer), F/T sensor |
| Max walking speed | | 1.8 km/h |

Figure 5.4 Specification of humanoid robot Gazelle

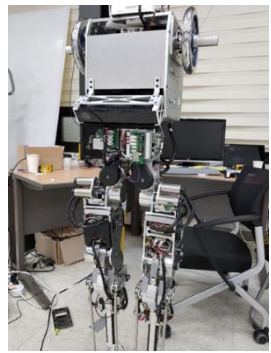
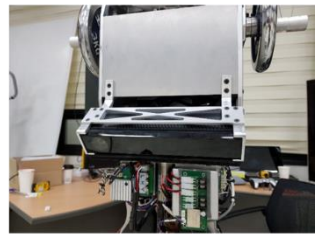

Figure 5.3 Camera mount to combine Gazelle and Kinect V2

Table 5.1 Kinematic constraints for real robot

| Kinematic Constraints | Value |
|---|---|
| Maximum step length | 0.30 m |
| Optimal step length | 0.20 m |
| Maximum step width | 0.35 m |
| Minimum step width | 0.18 m |
| Maximum step yaw | 15 deg |

In the real robot experiment, state estimation is done by kinematic information and IMU and FT sensor. It is because, visual odometry was too noisy than the simulation.

## 5.2. Real Robot Experiment Results

Experiment is taken in two scenarios. First one is a dynamic obstacle scenario. Robot is given to go to a goal in a straight direction, but suddenly a person gets in front of the robot. To get through the person, robot should solve local minimum problem in real time and finally get to the goal. Second one is a small object and large object scenario. Robot should get through a complex terrain map, which includes small obstacles and large obstacles. It solves the plan utilizing its adaptive action sets and finally reaches the goal while satisfying its feasibility condition.

1) Dynamic obstacles

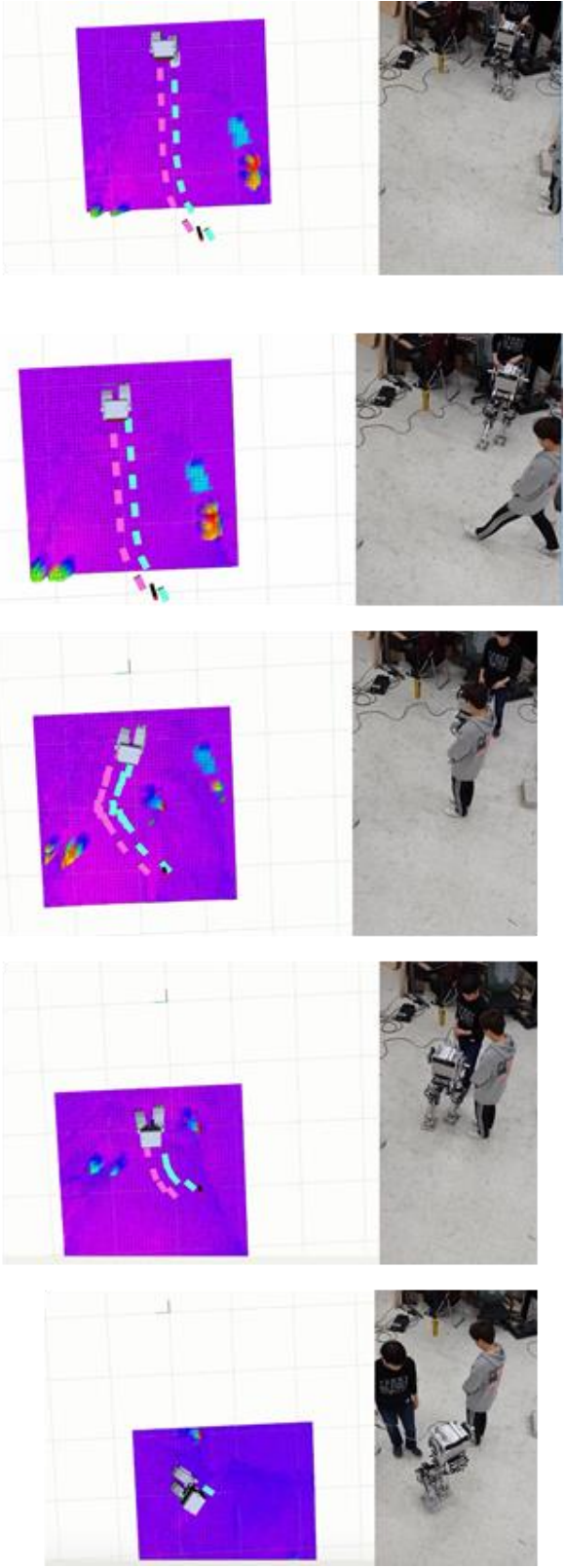

Figure 5.5 Dynamic obstacles experiment results in real robot.

2) Small and large obstacles.

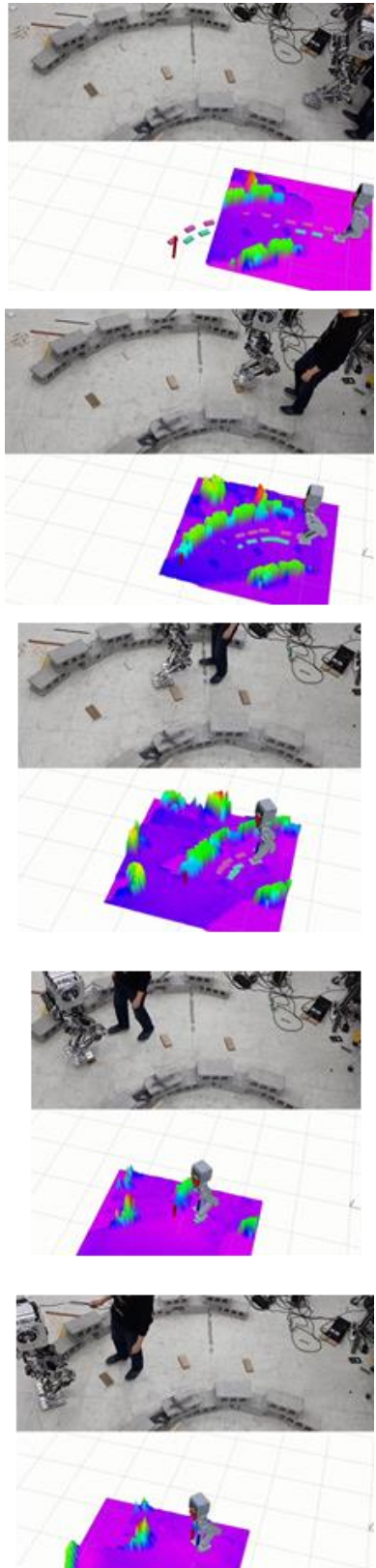

Figure 5.6 Small and large obstacles experiment results in real robot.

During the experiment, whole framework was run in real time to make it available to deal with dynamic obstacles. However, it had difficulties for environments which is out of FOV. This can be further improved by utilizing larger FOV sensors or, moreover, multiple sensors.

# Chapter 6. Conclusion

## 6.1. Conclusion

This thesis reviewed the history of humanoid navigation by footstep planning and proposed a framework that integrates adaptive action sets with human locomotion energy approximations by proposing a selection method by least COT. Furthermore, iteration time is reduced by setting heuristic cost function closer to the maximum limit for admissibility by considering angle difference in heuristic function. For the feasibility checking problem, dual level feasibility checking method is proposed, which utilizes footstep regions and ellipse regions. Furthermore, local minimum problem is reduced by penalizing heuristic functions about future obstacle collision. Finally, integrating framework for mapping, footstep planning, walking is introduced and verified not only in simulation, but also in real robot.

With these frameworks, humanoid navigation considering not only reaching the goal, but also human like energy efficiency is able. It is much desirable point of view in humanoid navigation, since it moves in a different way compared to the mobile robots. Also in a practical point of view, implementing real time navigation was able thanks to the various methods which minimizes computational effort while maximizing the essential effects. Although there exists more complex and desirable algorithms, real robot cannot be driven in real time with those methods.

Thus, author insists the framework introduced in this thesis is well integrated with computationally efficient methods while having philosophy that humanoid navigation should consider not only the distances, but also energy consumptions.

## 6.2. Future work

Currently used mapping algorithm is too slow due to its heavy computational effort and this makes the integrated map slow and wavy. In this reason, another mapping algorithm with less effort could replace the one used in this thesis.

Also, utilizing the environment data, real time obstacle feedback to walking algorithm may be possible. Currently footstep planner only makes desired footsteps. However, with the environment data, feedback can be done to the walking stabilizer to stabilize itself considering obstacles.

In addition, other precise collision checking methods could be considered which is not computationally heavy.

Finally, state estimator, which integrates visual odometry and kinematic odometry, can be adapted for more accurate state estimation.

# Appendix

A. Proof that $E_{angle}$ is proportional to $\theta^2$ for trajectory with trigonometric functions.

In order to have no discontinuous acceleration profile, frequently used trajectory plan is to utilize trigonometric functions. Also, since our system has constant step time, trajectory plan can be done as Equation (16). In the equation, $\theta_{des}$ is the desired yaw angle difference for a step and $T_{step}$ is the constant step time. One example of those trajectory is plotted as Figure A.1.

$$\begin{aligned}
position\ p(t) &= V_m \left\{ t - \frac{T_{step}}{2\pi} \sin\left(2\pi \frac{t}{T_{step}}\right) \right\} \\
velocity\ v(t) &= V_m \left\{ 1 - \cos\left(2\pi \frac{t}{T_{step}}\right) \right\} \\
acceleration\ a(t) &= 2\pi \frac{V_m}{T_{step}} \sin\left(2\pi \frac{t}{T_{step}}\right) \\
jerk\ j(t) &= 4\pi^2 \frac{V_m}{T_{step}^2} \cos\left(2\pi \frac{t}{T_{step}}\right) \\
&\ast V_m = \theta_{des} T_{step}
\end{aligned} \tag{16}$$

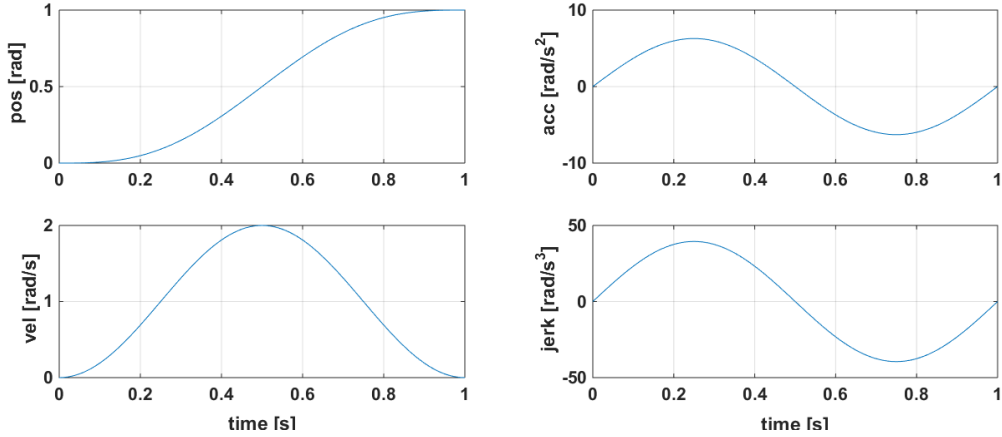

Figure A.1 Example of trajectory plan by trigonometric functions

If we assume that foot yaw movement can be expressed as a single load rotating in a yaw direction like Figure A.2. To make a movement that follows the trajectories like Equation (16), actuator needs to output torque as same as inertial torque regarding friction. Therefore, it consumes energy as Equation (17). After calculation, it is able to find out that it is proportional to $\theta_{des}^2$.

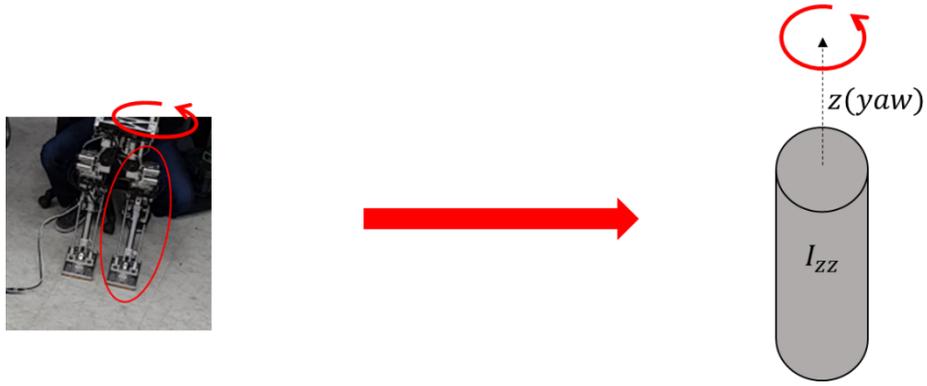

Figure A.2 Foot yaw movement assumed as a rotating single load

$$\int_0^{T_{step}} |I_{zz}a(t)v(t)|\ dt$$

$$= 2I_{zz}(2\pi T_{step}\theta_{des}^2) \int_0^{\frac{T_{step}}{2}} \sin\left(2\pi \frac{t}{T_{step}}\right) - \sin\left(2\pi \frac{t}{T_{step}}\right)\cos\left(2\pi \frac{t}{T_{step}}\right) dt$$

$$= 4I_{zz}\pi T_{step}\theta_{des}^2 \left( \int_0^{\frac{T_{step}}{2}} \sin\left(2\pi \frac{t}{T_{step}}\right)\ dt - \int_0^{\frac{T_{step}}{2}} \frac{\sin\left(4\pi \frac{t}{T_{step}}\right)}{2}\ dt \right) \quad (17)$$

$$= 4I_{zz}\pi T_{step}\theta_{des}^2 \left( -\frac{T_{step}}{2\pi}\cos\left(2\pi \frac{t}{T_{step}}\right) \Big|_0^{\frac{T_{step}}{2}} - 0 \right) = 4I_{zz}T_{step}^2\theta_{des}^2 \propto \theta_{des}^2$$

# Bibliography

[1] J. Kuffner, K. Nishiwaki, S. Kagami, M. Inaba, and H. Inoue, "Footstep planning among obstacles for biped robots," in Proc. IEEE/RSJ Int. Conf. on Intelligent Robots and Systems (IROS'01), 2001, pp. 500–505.

[2] J. Chestnutt, J. Kuffner, K. Nishiwaki, and S. Kagami, "Planning biped navigation strategies in complex environments," in Proc. IEEE-RAS/RSJ Int. Conf. on Humanoid Robots (Humanoids'03), Munich, Germany, Oct. 2003.

[3] J. Chestnutt, M. Lau, G. Cheung, J. Kuffner, J. Hodgins, and T. Kanade, "Footstep planning for the Honda ASIMO humanoid," in Proc. IEEE Int. Conf. Robot. Autom., Apr. 2005, pp. 629–634.

[4] J.-S. Gutmann, M. Fukuchi, and M. Fujita, "Real-time path planning for humanoid robot navigation," in Int. Joint Conference on Artificial Intelligence (IJCAI), Edinburgh, Scotland, 2005.

[5] N. Perrin, O. Stasse, L. Baudouin, F. Lamiraux, and E. Yoshida, "Fast humanoid robot collision-free footstep planning using swept volume approximations," IEEE Trans. Robot., vol. 28, no. 2, pp. 427–439, Apr. 2012.

[6] A. Hornung, A. Dornbush, M. Likhachev, and M. Bennewitz, "Anytime search-based footstep planning with suboptimality bounds," in Proc. 12th IEEE-RAS Int. Conf. Humanoid Robots, Nov. 2012, pp. 674–679.

[7] A. Stumpf, S. Kohlbrecher, D. C. Conner, and O. von Stryk, "Supervised footstep planing for humanoid robots in rough terrain tasks using a black box walking controller," in Humanoid Robots (Humanoids), 2014. 14th IEEE-RAS International Conference on. IEEE, 2014.

[8] P. Karkowski, S. Oßwald, and M. Bennewitz, "Real-time footstep planning in 3D environments," in IEEE-RAS 16th International Conference on Humanoid Robots (Humanoids), 2016, pp. 69–74.

[9] P. E. Hart, N. J. Nilsson, and B. Raphael, "A formal basis for the heuristic determination of minimum cost paths," IEEE Trans. Solid-Stste Circuits, vol. SSC-4, pp. 100–107, 1968.

[10] S. M. LaValle. Rapidly-exploring random trees: A new tool for path planning. TR 98-11, Computer Science Dept., Iowa State Univ. (http: //janouiec. cs. iastate.edu/papers/rrt .ps>, Oct. 1998.

[11] R. Dechter and J. Pearl. Generalized Best-first Search Strategies and the Optimality of A*. Journal of the ACM (JACM), 32(3):505–536, 1985.

[12] O. Kanoun, J.-P. Laumond, and E. Yoshida, "Planning foot placements for a humanoid robot: A problem of inverse kinematics," Int. Journal of Robotics Research (IJRR), 2011.

[13] R. Deits and R. Tedrake, "Footstep planning on uneven terrain with mixedinteger convex optimization," in Proc. 14th IEEE-RAS Int. Conf. Humanoid Robots, Nov. 2014, pp. 279–286.

[14] W. Huang, J. Kim, and C. Atkeson, "Energy-based optimal step planning for humanoids," in Proc. IEEE Int. Conf. Robot. Autom. May 2013, pp. 3124–3129.

[15] J. Kim, N. Pollard, and C. Atkeson, "Quadratic encoding of optimized humanoid walking," in Proc. 13th IEEE-RAS Int. Conf. Humanoid Robots, Oct. 2013, pp. 300–306.

[16] Martim Brandao, Kenji Hashimoto, et.al. "Footstep Planning for Slippery and Slanted Terrain Using Human-inspired Models", IEEE Transactions on Robotics, vol. 32, no. 4, 2016.

[17] J. Chestnutt, K. Nishiwaki, J. Kuffner, and S. Kagami, "An adaptive action model for legged navigation planning," in Proc. of the IEEE-RAS Int. Conf. on Humanoid Robots (Humanoids), 2007.

[18] Philipp Karkowski, Stefan Obwald, et. al. "Real-Time Footstep Planning in 3D Environments", in Proc. of the IEEE-RAS Int. Conf. on Humanoid Robots (Humanoids), 2016.

[19] J. Chestnutt and J. Kuffner, "A tiered planning strategy for biped navigation," in Proc. IEEE-RAS/RSJ Int. Conf. on Humanoid Robots (Humanoids'04), Santa Monica, California, Nov. 2004.

[20] A. Hornung and M. Bennewitz, "Adaptive Level-of-detail planning for efficient humanoid navigation," in Proc. IEEE Int. Conf. Robot. Autom., 2012, pp. 997–1002.

[21] A.-C. Hildebrandt, M. Klischat, D. Wahrmann, R. Wittmann, F. Sygulla, P. Seiwald, D. Rixen and T. Buschmann, Real-Time Path Planning in Unknown Environments for Bipedal Robots, IEEE Robotics and Automation Letters 2(4), 1856—-1863 (2017).

[22] E. Yoshida, C. Esteves, I. Belousov, J.-P. Laumond, T. Sakaguchi, and K. Yokoi, "Planning 3-D collision-free dynamic robotic motion through iterative reshaping," IEEE Trans. Robot., vol. 24, no. 5, pp. 1186–1198, Oct. 2008.

[23] D. Wahrmann, A.-C. Hildebrandt, R. Wittmann, F. Sygulla, D. Rixen, and T. Buschmann, "Fast Object Approximation for Real-Time 3D Obstacle Avoidance with Biped Robots," in IEEE International Conference on Advanced Intelligent Mechatronics., 2016.

[24] Lim, Jeongsoo, et al. "Robot system of DRC-HUBO+ and control strategy of team KAIST in DARPA robotics challenge finals." Journal of Field Robotics 34.4 (2017): 802-829.

[25] ROS [Online] https://www.ros.org

[26] Gazebo [Online] https://gazebosim.org

[27] "Rtabmap", M. Labbe and F. Michaud, "Online global loop closure detection for large-scale multi-session graph-based SLAM," in Proc. IEEE/RSJ Int. Conf. Intell. Robot. Syst., Sep. 2014, pp. 2661–2666.

[28] Fankhauser, Péter, et al. "Robot-centric elevation mapping with uncertainty estimates." Mobile Service Robotics. 2014. 433-440.

[29] Fankhauser, Péter, and Marco Hutter. "A universal grid map library: Implementation and use case for rough terrain navigation." Robot Operating System (ROS). Springer, Cham, 2016. 99-120.

[30] R. B. Rusu and S. Cousins, "3D is here: Point Cloud Library (PCL)," presented at the IEEE Int. Conf. Robot. Autom., Shanghai, China, May 9–13, 2011.

[31] A. Hornung, K. M. Wurm, M. Bennewitz, C. Stachniss, and W. Burgard, "OctoMap: An efficient probabilistic 3D mapping framework based on octrees," Auton. Robots, pp. 189–206, 2013.

[32] Matt Zucker. "Quadruped footstep planning with A*". E28 Mobile Robotics. 2016

[33] Fankhauser, Péter, et al. "Kinect v2 for mobile robot navigation: Evaluation and modeling." 2015 International Conference on Advanced Robotics (ICAR). IEEE, 2015.

[34] Microsoft, Kinect for Windows features,"http://www.microsoft.com/en-us/kinectforwindows/meetkinect/features.aspx" January, 2015.

[35] Kajita, Shuuji, et al. "Biped walking pattern generation by using preview control of zero-moment point." *ICRA*. Vol. 3. 2003.

[36] Jung, Taejin, et al. "Development of the Humanoid Disaster Response Platform DRC-HUBO+." IEEE Transactions on Robotics 34.1 (2018): 1-17.

# Acknowledgement in Korean

    길고도 짧은 2년의 석사과정 기간 동안 혼자 만의 싸움이 아니라, 곁에 있어준 많은 사람들이 있었기에 끝맺음을 지을 수 있었던 것 같습니다. 이 글을 통해 그 분들께 감사 드리는 마음을 전하고자 합니다.

    먼저, 사랑하는 가족들에게 고마움을 전하고 싶습니다. 아버지, 아버지는 저에게 하나의 큰 등대이자 동료입니다. 매일 연구자의 자세를 솔선수범하여 보여주시기 때문에, 저도 항상 그 모습을 보며 존경심을 가지고 연구자의 길을 걸어갈 수 있는 것 같습니다. 많이 의견 충돌이 일어나기도 하지만, 결국에는 아버지의 말씀들이 저에게 가장 큰 힘이 되었습니다. 어머니, 어머니의 무한한 사랑 덕분에 저는 마음 한 켠에 항상 따뜻함이 있을 수 있던 것 같습니다. 예전에는 잘 몰랐지만, 요즘 들어서 많이 웃고 행복해 하시는 모습들을 볼 때면 너무 사랑스럽습니다. 누나, 감정적으로 힘든 일들이 있을 때마다 많이 기댔었는데 그 때마다 너무 좋은 말들 해줘서 항상 힘이 났어, 누나가 정말 이 사회에 필요하고 어려운 일을 하고 있기 때문에 항상 응원하고 또 내가 도움이 될 수 있는 부분들에서는 많이 도우려고 노력할게.

    다음으로, 항상 묵묵히 저를 지지해주시고 응원해주신 교수님께 감사드립니다. 제가 연구에 막혀 전전긍긍할 때에, 넓은 시야로 중요한 점들을 먼저 바라보라고 조언 해주신 덕분에 제 연구의 결과물을 볼 수 있었던 것 같습니다. 또한, 바쁘신 와중에도 저의 발표들 들어주시고 진심으로 조언 해주신 김진환 교수님, 김아영 교수님께도 감사드립니다. 연구실의 활기를 불어넣어 주시고 지나가면서 인사할 때마다 반갑게 받아주시는 박현섭 교수님도 감사드립니다.

    다음으로, 저의 연구와 연구실 생활이 풍요롭고 또 즐겁게 도와주신 휴보랩 구성원들에게 항상 감사하다는 말을 전하고 싶습니다. 김민수 선배님, 하드웨어를 따로 다루지 않아서 직접적인 도움을 많이 받지는 못하였지만 세미나시간마다 조언 해주시고 또 농구할 때 같이 즐겼던 기억들이 있습니다. 연구실 생활 적응하는 데 도움 주셔서 감사합니다. 태진이형, 많은 접점은 없었지만 형님을 보면서 운동을 해야겠다는 생각을 가지게 되었고, 또 형님의 설계 센스와 속도에 항상 존경심을 가지게 되었습니다. 최근에 얘기 나누었던 딥러닝 연구 많은 도움은 드리지 못했지만 좋은 결과 있으셨으면 좋겠습니다. 현민이형, 형 덕분에 제 주제에 대해서 고민해볼 기회가 생겼었습니다, 항상 고마운 마음 가지고 있습니다. 형이 연구할 때 DRC-HUBO+ 워킹 부분 도와주셔서 제 연구가 결과물로서 보여질 수 있었습니다. 제가 힘들어할 때마다 차분하게 웃으면서 얘기 해주셔서 큰 힘이 되었습니다. 옥기형, 형의 코멘트들이 제 연구에 가장 큰 도움이 되었습니다. 발표 준비 때도 가장 적극적으로 들어주시고 조언해 주셔서 너무 감사했습니다. 배드민턴을 제가 잘 못 쳐서 같이 많이 못 쳐본 게 아쉽고, 형이 항상 나눠주는 과자들 너무 감사했습니다. 효인이형, 항

상 바쁘셔서 얘기를 많이 못 나눈 게 너무 아쉽지만, 가끔 질문할 때마다 정말 친절하게 자세히 알려주셔서 많은 도움이 되었습니다. 특히 제 연구에서 중요한 부분 중 하나인, 상태 추정 알고리즘 부분에 많은 도움이 있었기에 제 연구가 결과물로서 연결될 수 있었던 것 같습니다, 감사드립니다. 효빈이형, 연구 후반 실제 로봇에 적용할 때에 정말 많은 시간을 공유하며 같이 협업하였었는데, 항상 현명하신 판단과 리더십으로 저를 잘 이끌어주셔서 감사합니다. 연구 외적으로도 항상 재미있는 얘기 많이 해주시고 같이 농구도 하고, 정말 즐거운 일들이 가득하게 해주셔서 감사합니다. 재성이형, 연구실에서 가장 많이 대화하고 자전거도 타면서 많은 추억을 가질 수 있었습니다. 제가 힘들 때 얘기 많이 들어주시고, 연구로 지쳐있을 때 조용히 음료수 하나 가져다 주실 때 정말 감동 많이 받았습니다. 앞으로도 까불까불하고 부족하지만 항상 형을 존경하는 후배인 저랑 잘 지내 주시면 감사하겠습니다. 강규형, 실장 하실 때부터 카리스마 넘치는 리더십으로 좋은 조언들 많이 해주시고, 또 한편으로는 위트 넘치는 유머들로 저를 웃게 만들어주셔서 감사합니다. 발표 준비할 때도 정말 부족한 부분들 끝까지 잘 들어주시고 최대한 발전할 수 있는 방향으로 조언해 주시려는 모습들을 보면서 정말 감동이었고, 연구에 집중하시는 모습들이 저에게 많은 동기부여가 되었기에 감사드립니다. 부연이형, 여리고 마음이 너무 따뜻한 제가 너무 좋아하는 형, 항상 힘들어하실 때마다 많은 도움이 되지 못한 것 같아서 너무 죄송한 마음이 많이 듭니다. 되돌아보면 형이 해주시는 음식들은 항상 정말 너무 맛있었어요, 같이 자전거 타러 가면서 많은 추억들도 쌓이고 또 장난스런 얘기들 할 때마다 웃으시는 모습이 너무 귀여우셔서 연구실에서 제가 행복할 수 있었던 것 같습니다. 앞으로도 같이 연구실 생활 하면서 많은 추억을 쌓고 싶고 잘 부탁드리겠습니다. 성우형, 차분한 것 같으면서도 장난기 많으신 성우형, 헬스 같이 하면서 장난 칠 때마다 잘 받아주셔서 너무 감사해요. 형이랑 같이 해피하우스 살 때에도 오고 가며 좋은 말씀 많이 해주셔서 힘이 됐었는데, 그 이후로도 항상 차분하게 제 얘기 잘 들어주시고 또 공감해주시기에 연구실 생활 할 때에 행복했습니다. 내일도 또 같이 운동하러 가시죠. 승우형, 다정하면서도 친근하고 또 장난도 많이 치는 형, 형이랑 같이 헬스 하고 노래방 갈 때마다 웃음이 끊이질 않는 거 같아요. 연구실에서 가장 먼저 나오고 가장 늦게 가시는 성실함이 항상 존경스럽고, 어른스럽고 또 어쩔 때는 아이 같은 모습들이 저에게는 정말 편안하게 다가왔던 거 같아요. 그래서 고민 상담을 형에게 정말 많이 하게 된 것 같은데, 그 때마다 진지함과 유머를 정말 잘 섞어서 얘기해 주셔서 항상 너무 힘이 났어요. 앞으로도 같이 많은 얘기하면서 좋은 선후배 이자 친구가 됐으면 좋겠어요, 감사합니다 형님. 동현이형, 형의 연구를 도와드리고 또 얘기 나누면서 연구에 대한 관점이 많이 잡히게 된 것 같아요. 특히 자료를 정말 잘 정리하고, 기록을 꼭 남기는 형의 습관들이 저는 너무 존경스러웠어요. 연구 외적으로도 많이 친하게 지내 주셔서 너무 고마웠고, 지금은 다른 연구실에 가셨지만 그 곳에서도 항상 좋은 연구와 행복한 일만 가득하셨으면 좋겠어요, 감사합니다. 다은누나, 힘들어 하실 때 많은 도움이 되지 못한 것 같아서 너무 죄송한 마음이 앞서네요, 그래도 연구 잘 마무리 하시고 졸업하실 때 정말 너무 멋있고 자랑스러웠습니다. 졸업하신 이후에도 항상 좋은 일만 가득하시고 행복하시길 바래요, 감사합니다. 동기 승훈이형, 내가 가장 존

경하는 사람 중 한명인 승훈이형, 정말 누구보다 자기 연구를 사랑하고 또 열심히 하는 모습들을 보면서 많이 배우게 된 것 같아, 형 덕분에 힘들 때마다 많이 의지도 됐고 또 많은 동기부여도 생겼어. 대학교 때 동아리 회장단 같이 할 때부터 서로 정말 장단점을 잘 보완해주는 존재인 것 같아서, 형을 만나게 된 게 나는 내 인생에 큰 행운이라고 생각해, 앞으로도 같이 연구할 때 서로 많은 도움이 되는 동료이자 또 연구 외로는 좋은 친구로써 잘 지내자. 의욱이형, 후배로 들어왔지만 실질적으로는 선배인 의욱이형, 형이랑 같이 축구 할 때마다 너무 재미있어요. 실전 경험이 풍부한 만큼 앞으로 좋은 연구 할 수 있을 거라고 생각해요. 앞으로도 재미있게 같이 추억 만들었으면 좋겠습니다. 종훈이, 못난 모습만 많이 보여줘서 많이 미안하고, 선배로써 도움이 많이 되지 못한 거 같아서 또 미안해. 툴툴대면서도 자상한 모습들이 항상 멋있고, 자기 할 일 꾸준하게 열심히 하는 게 대단하다고 생각해. 앞으로도 부족한 형 많이 챙겨주면 고마울 거 같아. 순표, 너를 보고 있으면 저절로 웃음이 나오는 거 같아. 그만큼 긍정적이고 또 공부할 때는 누구보다 열심히 그리고 창의적인 모습들을 보면서 항상 대단하다고 생각해. 관심 분야가 비슷한 부분이 있는 만큼 앞으로 같이 협력해서 연구 하면 좋을 거 같아 힘내 보자.

KI연구원 분들에게도 많은 감사드립니다. 항상 보고서 관련하여 궂은 일들을 맡아서 해주시는 양현대 박사님, 시연 도맡아서 해주시고 지나갈 때마다 재미있는 얘기 많이 해주시는 진용이형, 처음 같이 footstep planning을 협업 했었고 관련 내용들에 대해서 토론도 많이 한 문영이형, 시연들 많이 해주시고 노래도 정말 잘하시는 유진누나, 나중에 들어오셨지만 Gazebo와 ROS 관련해서 같이 얘기 많이 나눈 새힘누나, 정말 따뜻한 마음으로 친근하게 다가와 주시고 보듬어 주시는 용정이형, 서류작업들을 도맡아서 정말 잘 해주시는 박유나 선생님, 민희누나 모두 정말 감사드립니다.

마지막으로, 사적으로 많은 힘이 되어준 친구들과 친척분들도 모두 너무 감사드리며, 짧고도 긴 석사 생활을 하며 배운 것들을 토대로 앞으로 박사과정에 진학한 이후에도 인류에 이바지 하는 좋은 연구를 하는 연구자가 될 것을 다짐합니다.